\documentclass[11pt]{article}

\usepackage[a4paper,margin=1in]{geometry}
\usepackage{amsmath,amssymb}
\usepackage{graphicx}
\usepackage{booktabs}
\usepackage{float}
\usepackage[numbers,sort&compress]{natbib}
\usepackage{hyperref}
\usepackage{orcidlink}
\usepackage{xcolor}
\usepackage{tikz}
\usepackage{subcaption}
\usetikzlibrary{positioning,arrows.meta,shapes.multipart,fit,backgrounds}

\title{Before AI Agents Negotiate for Us: A Competitive Tournament of Rule-Based, Search, Optimization, Learned, and Language-Model Negotiators Under Scarcity}

\author{%
Eduardo de Avila-Armenta$^{1,2}$\,\orcidlink{0000-0003-3461-7722}, Ra\'ul Alejandro Aquino-Ibarra$^{2}$, \\
Jorge Octavio Nicol\'as-D\'iaz$^{2}$, Umberto Leon-Dominguez$^{2,3}$\,\orcidlink{0000-0001-5170-2553}\thanks{Corresponding author. Email: \texttt{umberto.leon@udem.edu}} \\[4pt]
\small $^{1}$Tecnol\'ogico de Monterrey, School of Engineering and Sciences, \\
\small Av. Eugenio Garza Sada 2501 Sur, Tecnol\'ogico, 64849 Monterrey, N.L., Mexico \\
\small $^{2}$AI Lab, Freight Technologies Inc. \\
\small $^{3}$School of Psychology, University of Monterrey, Monterrey, Mexico
}

\date{\today}

\begin{document}

\maketitle

\begin{abstract}
Automated bilateral negotiation underlies two-sided matching markets such as freight brokering, where many shippers and carriers negotiate price before a matching mechanism decides who is paired with whom.
Prior strategy comparisons typically use isolated one-on-one duels or a single fixed market, so they cannot show how a strategy behaves against several live rivals under a changing balance of supply and demand.
We built nine negotiation architectures spanning rule-based, search-based, optimization-based, imitation-learning, and reinforcement-learning families, and ran all nine simultaneously on both sides of a Gale-Shapley matching market under a balanced condition and two scarcity conditions rotated across all 36 exclusion pairs, with every pairwise comparison Holm-Bonferroni corrected.
Eight of the nine architectures share one normalized \emph{negotiation distance} -- the position of a price between the agent's worst acceptable deal (0) and its ambitious target (1), i.e.\ how close the table is to a deal the agent wants -- as their common scale for valuing, optimizing, or learning offers.
In the balanced carrier market, an imitation-learned engine (ML2), a reinforcement-learning policy trained only on isolated negotiations (RL), and an expected-utility optimization engine (OPT) formed a statistically tied top group, and no architecture led in every condition and role.
RL beat all eight carrier rivals when carriers were scarce, but its match rate collapsed from 62\% to 16\% when shippers were scarce and it had to compete with the other carrier agents for demand.
A second reinforcement-learning policy trained against rival agents (RL2) kept its match rate above 91\% in every condition, yet lost on realized payoff to all eight shipper-side rivals in two of the three conditions, traced to a confirmed train/deployment feature mismatch.
We also tested nine large language models (LLMs), from 1 to 72 billion parameters, under the same protocol: they lost significantly to the two strongest engines (OPT2 and ML2) in 32 of 36 comparisons and won none.
The strongest model, \texttt{llama3.3:70b}, posted the highest raw profit of all twelve shipper-side participants without reaching significance, while \texttt{deepseek-r1:70b} collapsed to a 2.5\% match rate because its responses hit the 90-second timeout on our single-GPU workstation.
These results show that isolated or single-condition evaluation can hide failure modes that a many-sided, scarcity-varying, statistically corrected tournament exposes directly.

\end{abstract}

\section{Introduction}
\label{sec:intro}

Two-sided matching markets in which many buyers and many sellers must both agree on a price and be paired with a counterpart -- freight brokering, ride-hailing dispatch, labor and housing markets -- are negotiated today by human brokers, by software agents, or by both, and increasingly depend on autonomous negotiation agents (this study evaluates software agents only, with no human participants) that can reach a deal quickly, on favorable terms, while remaining attractive enough to actually be the counterpart someone else chooses.
A negotiation agent in such a market therefore faces two linked problems at once: winning a good price at the bilateral table, and winning the competition for a match once every table's outcome is fed into a stable-matching mechanism \citep{gale1962} that may discard an agreed deal in favor of a better one found elsewhere.

Work on automated negotiation strategies has produced a wide range of architectures -- hand-written concession rules, value-function-guided heuristics, lookahead search over sampled opponent models, continuous expected-utility optimization, strategies that imitate a stronger policy's offers, and reinforcement-learned policies trained end to end \citep{florijn2024,cowling2012,schulman2017} -- but these architectures are typically compared in one of two limited ways: isolated one-on-one duels against a small number of fixed opponents, or a single population competing under one fixed market condition.
Neither setting can reveal how an architecture behaves once it must negotiate against several rival architectures simultaneously, nor how its behavior changes as the balance between supply and demand shifts from a balanced market toward scarcity on one side.
This gap matters because the matching stage can decouple negotiation skill from final outcome entirely: an architecture that reaches good prices in isolation can still fail to be matched at all if it is rarely the counterpart a scarce resource prefers, and a reinforcement-learned policy trained only against a fixed, non-adapting opponent has no guarantee that what it learned survives contact with genuine competitive pressure from rival architectures of its own side.

Several of this paper's hand-designed architectures -- the rule-based, search-based, and optimization-based engines, whose decision logic is written explicitly rather than learned from data -- encode specific findings from the negotiation-psychology literature rather than arbitrary heuristics: an ambitious opening anchor calibrated to the observed spread between the parties' positions \citep{vanpoucke2002,petrowsky2025}, a geometrically decreasing concession schedule \citep{tey2021}, every offer framed against the counterpart's own reference point \citep{kristensen1997}, and a portfolio of concurrent tables treated as leverage in the same way holding multiple outside alternatives distorts a negotiator's perceived bargaining zone \citep{schaerer2016,florijn2024} -- design choices this paper's tournament lets compete against each other directly, rather than evaluating any one of them in the isolated, single-negotiation settings the underlying behavioral studies used.
Large language models are a natural further architecture to include in such a tournament: a concurrent study found that an LLM prompted to select game moves directly loses to an agent built around an explicit, LLM-synthesized world model combined with search, in the large majority of the games tested \citep{lehrach2025} -- a direct-policy-versus-model-and-search pattern this paper tests again in a different domain (Section~\ref{sec:results-llm}).

This paper reports a controlled tournament designed to surface exactly these effects.
We implemented nine negotiation-engine architectures, spanning rule-based, value-function-augmented, search-based, continuous-optimization, imitation-learning, and reinforcement-learning designs, on both the shipper and carrier side of a bilateral freight-price negotiation, and ran all nine simultaneously -- never in isolated duels -- under three market conditions: a balanced market with all nine architectures present on both sides, and two scarcity conditions in which one side's roster is reduced from nine to seven architectures, with the excluded pair rotated systematically across all $\binom{9}{2}=36$ possible exclusions so that no architecture is systematically favored or penalized by the experimental design itself.
Every pairwise comparison between two architectures under a given market condition and role is tested with a paired Wilcoxon signed-rank test on same-day outcomes and corrected for the resulting 36 simultaneous comparisons with the Holm-Bonferroni procedure, and every architecture's own outcome is summarized with a percentile bootstrap confidence interval rather than a normal-approximation interval, since margin and profit outcomes in this market are not symmetrically distributed.

The results show that formulation sophistication alone does not guarantee good outcomes under competition: a reinforcement-learned policy trained only on isolated negotiations suffers a severe match-rate collapse once rival agents (the other architectures of its own side, all of them software agents) compete with it for scarce demand, while the same architecture trained against rival agents avoids that collapse but is instead undone by a confirmed mismatch between the feature values it saw during training and the feature values it receives at deployment time, a finding that would not surface in an isolated duel.
We report both findings directly, alongside an extension of the same tournament to nine large-language-model-based negotiation agents, including a qualitative look at individual negotiation trajectories reconstructed from the same call logs.
Section~\ref{sec:related} positions this tournament against prior automated-negotiation platforms, learned dialogue agents, and recent LLM-negotiation benchmarks; Section~\ref{sec:methods} describes the negotiation protocol, market model, matching mechanism, the nine architectures, the experimental design, and the statistical analysis; Section~\ref{sec:results} reports the tournament's results; Section~\ref{sec:discussion} discusses their implications, compares them against that prior work, and states limitations; Section~\ref{sec:conclusion} concludes.

\section{Related Work}
\label{sec:related}

Automated negotiation has been studied through two largely separate ecosystems, each with its own standard platform and evaluation protocol, and a third, more recent line specific to large language models.

The first is the GENIUS/ANAC ecosystem \citep{lin2014genius}: a shared Java platform for specifying negotiation domains (a fixed set of issues and one private linear-additive utility function per party) and running bilateral or small multilateral sessions between submitted agents, hosted yearly as the International Automated Negotiating Agents Competition since 2010.
GENIUS/ANAC's domains and its published yearly rankings are the closest thing automated negotiation has to a shared benchmark, but every session is still an isolated duel (or a small fixed multilateral group) over one utility function, with no matching stage that can discard an agreed deal and no systematic variation in how many rivals of the same side are present.
\citet{dejonge2022micro} shows this isolation has a concrete cost: MiCRO, a strategy with no opponent modeling or learning at all, outperforms the strongest submitted agents across four ANAC editions simply by exploiting regularities of the classic linear-additive domains themselves, a result later given a full theoretical treatment \citep{dejonge2024micro}.
That finding argues the classic ANAC domains are too simple on the \emph{utility-function} axis; this paper's tournament instead varies difficulty on a different axis entirely, the \emph{market-structure} axis (how many rivals are present and how scarce the opposite side is), while keeping each individual bilateral exchange comparatively simple.

The second ecosystem is end-to-end learned negotiation dialogue, seeded by \citet{lewis2017}'s "Deal or No Deal" task (a multi-issue bargaining dialogue with hidden per-party rewards, solved with a sequence-to-sequence policy trained by self-play and rollouts) and extended by \citet{he2018}'s CraigslistBargain, which decouples a coarse dialogue-act strategy from natural-language generation to make the learned strategy itself inspectable.
Both trained and evaluated their agents in isolated bilateral dialogues against a fixed human or self-play distribution, never against a population of rival negotiators competing for the same scarce counterparts -- the same isolation this paper's RL architecture (Section~\ref{sec:architectures}) exhibits by design, and whose cost under real competitive pressure Section~\ref{sec:results-collapse} measures directly.

The third, most recent line evaluates large language models themselves as negotiators, outside any hand-designed architecture.
\citet{abdelnabi2023} scores LLM agents in multi-issue negotiation games with tunable difficulty and finds a large capability gap between model generations, though a 2026 reproduction \citep{pollo2026reproduction} finds that ranking models on this benchmark is less objective than it first appears, cautioning against reading any single LLM-negotiation leaderboard as settled.
\citet{termsbench2026} and \citet{oh2026merit} both push past a single deal-rate or win-rate summary, diagnosing surplus extraction, belief calibration, and strategy quality separately -- the same spirit as this paper's own decomposition of the LLM comparison into match rate, realized margin/profit, and the qualitative trajectory case studies of Section~\ref{sec:results-llm-trajectories}, rather than a single pass/fail number per model.
None of these three LLM-negotiation benchmarks places its agents inside a populated market with a real matching mechanism or a rival roster of hand-designed and learned (non-LLM) architectures; \citet{lehrach2025}'s finding that a direct LLM policy loses to an explicit model-and-search agent in most games tested is, to our knowledge, the closest prior result to this paper's own finding that language models lose to this tournament's two strongest engines in the large majority of comparisons (Section~\ref{sec:results-llm}), but it is measured on a disjoint set of classic games, not a negotiation market.

Across all three lines, no prior evaluation places rule-based, search-based, optimization-based, imitation-learned, reinforcement-learned, and LLM-based negotiation agents under one shared, statistically corrected protocol, inside a populated market with a real matching mechanism and systematically rotated scarcity.
That combination -- not a harder utility function, and not a single additional agent family, but the market structure and the shared cross-paradigm comparison together -- is this paper's point of departure from each of the three lines above, and is revisited directly against each of them in Section~\ref{sec:discussion-sota}.

\section{Methods}
\label{sec:methods}

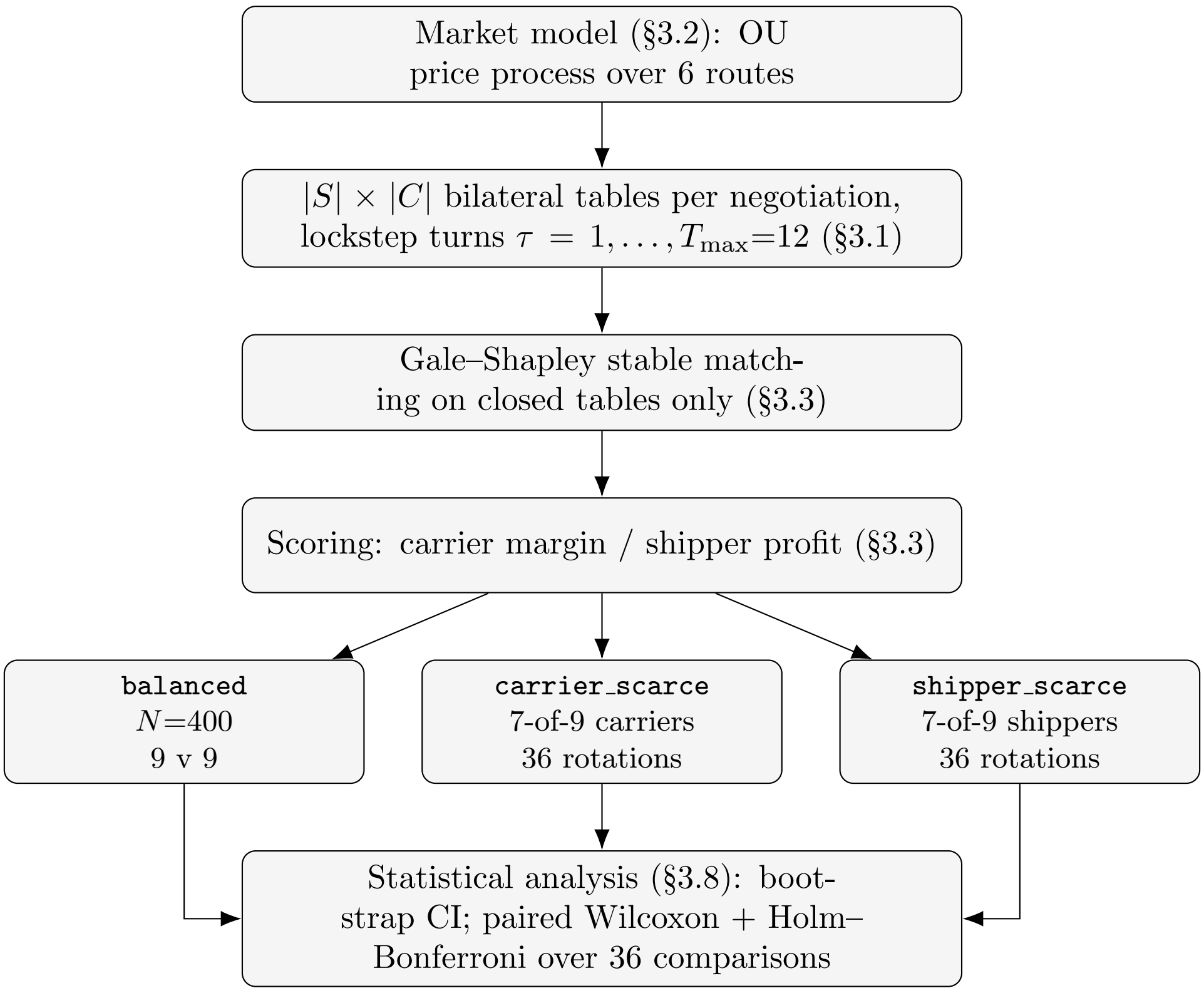
\begin{figure}[H]
\centering
\begin{tikzpicture}[
  node distance=7mm and 6mm,
  box/.style={draw, rectangle, rounded corners, align=center, text width=7.2cm, minimum width=7.6cm, minimum height=1.0cm, font=\small, fill=gray!8},
  cond/.style={draw, rectangle, rounded corners, align=center, text width=3.5cm, minimum width=3.8cm, minimum height=1.3cm, font=\footnotesize, fill=gray!8},
  arr/.style={-{Latex[length=2.2mm]}}
]
\node[box] (market) {Market model (\S\ref{sec:market}): OU price process over 6 routes};
\node[box, below=of market] (tables) {$|S|\times|C|$ bilateral tables per negotiation,\\ lockstep turns $\tau=1,\dots,T_{\max}{=}12$ (\S\ref{sec:protocol})};
\node[box, below=of tables] (matching) {Gale--Shapley stable matching on closed tables only (\S\ref{sec:matching})};
\node[box, below=of matching] (scoring) {Scoring: carrier margin / shipper profit (\S\ref{sec:matching})};
\node[cond, below=of scoring] (carrierscarce) {\texttt{carrier\_scarce}\\ 7-of-9 carriers\\ 36 rotations};
\node[cond, left=of carrierscarce] (balanced) {\texttt{balanced}\\ $N{=}400$\\ 9 v 9};
\node[cond, right=of carrierscarce] (shipperscarce) {\texttt{shipper\_scarce}\\ 7-of-9 shippers\\ 36 rotations};
\node[box, below=of carrierscarce] (stats) {Statistical analysis (\S\ref{sec:stats}): bootstrap CI; paired Wilcoxon $+$ Holm--Bonferroni over 36 comparisons};
\draw[arr] (market) -- (tables);
\draw[arr] (tables) -- (matching);
\draw[arr] (matching) -- (scoring);
\draw[arr] (scoring) -- (balanced);
\draw[arr] (scoring) -- (carrierscarce);
\draw[arr] (scoring) -- (shipperscarce);
\draw[arr] (balanced) |- (stats);
\draw[arr] (carrierscarce) -- (stats);
\draw[arr] (shipperscarce) |- (stats);
\end{tikzpicture}
\caption{Full experimental pipeline, from the market model through the three market conditions (Section~\ref{sec:design}) to the statistical analysis (Section~\ref{sec:stats}). $S$ is the set of shippers and $C$ the set of carriers in one negotiation, so $|S|\times|C|$ is the number of shipper-carrier tables negotiated in parallel; OU is the Ornstein-Uhlenbeck price process (Eq.~\eqref{eq:ou}).
Every negotiation runs independently through the top four stages; which of the three condition boxes it belongs to only changes how many architectures are present on which side.}
\label{fig:pipeline}
\end{figure}

Figure~\ref{fig:pipeline} summarizes the full pipeline described in this section, from the market model through the three market conditions of Section~\ref{sec:design} to the statistical analysis of Section~\ref{sec:stats}.

\subsection{Negotiation protocol}
\label{sec:protocol}

One negotiation instantiates a set of shippers $S$ and a set of carriers $C$, who interact over every one of the $|S|\times|C|$ bilateral tables simultaneously, one table per shipper-carrier pair.
Every table runs on the same global turn clock $\tau=1,\dots,T_{\max}$ with $T_{\max}=12$: shippers act on odd $\tau$ and carriers act on even $\tau$, so each side has exactly six decision opportunities per negotiation.
At each decision, an engine receives a table state containing the full turn history of that table (every offer either side has made), the counterpart's current offer if any, and a snapshot of the agent's \emph{other} live tables in the same negotiation (whether each is closed, in impasse, or still open, and its latest offer) -- this portfolio view is the only information an architecture has about its own concurrent negotiating load, and it never sees any information about a rival architecture's tables.
A carrier's legal offer range (its ``hard band'') is
\begin{equation}
\label{eq:carrier-band}
\begin{aligned}
RP_t &= \bar\mu - \tfrac{1}{2}\sigma, \\
\mathrm{Tgt}_t &= \bar\mu + \tfrac{1}{2}\sigma,
\end{aligned}
\end{equation}
where $\bar\mu$ and $\sigma$ are the route's true reference-price mean and standard deviation -- synthetic values drawn by the market model of Section~\ref{sec:market}, not taken from real freight data, and, like every price, rate, margin, and profit in this paper, expressed in US dollars (USD) -- and the subscript $t$ denotes this table (held fixed for the whole negotiation), not the turn counter $\tau$ used below; every carrier offer $x$ must satisfy $x\in[RP_t,\mathrm{Tgt}_t]$ and must never exceed the carrier's own previous offer on that table (monotonically non-increasing concessions).
A shipper's legal offer range is bounded above by its hard reservation price and anchored below by an ambitious target,
\begin{equation}
\label{eq:shipper-band}
\begin{aligned}
RP &= p_{\text{sale}} - c_{\text{goods}}, \\
\mathrm{Tgt} &= \widehat{RP}_{\text{est}} - \tfrac{1}{2}\sigma + k\,\sigma_\varepsilon,
\end{aligned}
\end{equation}
where $p_{\text{sale}}$ is the shipper's sale price for the shipment, $c_{\text{goods}}$ is its goods cost, $\widehat{RP}_{\text{est}}$ is the shipper's own noisy estimate of the route's true reference price (Section~\ref{sec:market}), $\sigma_\varepsilon=0.3\,\sigma$ is that estimate's noise scale, and $k$ is a per-engine cushion constant; every shipper offer must satisfy $x\le RP$ and must never fall below the shipper's own previous offer (monotonically non-decreasing concessions).
An offer violating either the band or the monotonicity constraint is an illegal move and is treated as an automatic forfeit of that table, exactly like an explicit reject; an accept is only legal when something is actually on the table and that offer already satisfies the acceptor's own band.
If a table reaches $\tau=T_{\max}$ without a deal, it ends in impasse and never enters the matching stage.

\subsection{Market model}
\label{sec:market}

All prices in this market are synthetic and in USD.
Six routes are drawn once per round -- the balanced condition's 400 negotiations form one round, and each of a scarce condition's 36 sub-rounds of 12 negotiations is its own round (Section~\ref{sec:design}) -- from
\begin{equation}
\label{eq:route}
\begin{aligned}
\mu_r &\sim \mathrm{Unif}(800, 3000), \\
\sigma_r &= 0.15\,\mu_r, \qquad r=1,\dots,6,
\end{aligned}
\end{equation}
and in every negotiation of that round each shipper is assigned one of the six routes at random; $\mu_r$ and $\sigma_r$ instantiate $\bar\mu$ and $\sigma$ in Eq.~\eqref{eq:carrier-band} for every table on that route.
The shipper does not observe $\mu_r$ directly: it only observes a noisy estimate
\begin{equation}
\label{eq:shipper-noise}
\begin{aligned}
\widehat{RP}_{\text{est}} &= \mu_r + \epsilon, \\
\epsilon &\sim \mathcal{N}\!\left(0,\ (0.3\,\sigma_r)^2\right),
\end{aligned}
\end{equation}
which is the one deliberate information asymmetry in the market: the carrier's true band is exact, the shipper's belief about it is not.
Sale price and goods cost are drawn once per negotiation index $n=1,\dots,N$ from an independent, discretized, log-space Ornstein-Uhlenbeck process for each quantity,
\begin{equation}
\label{eq:ou}
\begin{aligned}
\log P_n &= \log P_{n-1} + \kappa\left(\log P_0 - \log P_{n-1}\right) + \eta_n, \\
\eta_n &\sim \mathcal{N}(0,\sigma_{OU}^2),
\end{aligned}
\end{equation}
with $\kappa=0.2$ and $\sigma_{OU}=0.12$, clamped to a fixed $[\text{min},\text{max}]$ range after each step ($[1500,15000]$ for sale price, base $5000$; $[500,8000]$ for goods cost, base $2000$), so that both prices mean-revert around their base value rather than drifting without bound.

\subsection{Matching and scoring}
\label{sec:matching}

Once every table in a negotiation has closed or reached impasse, the set of \emph{closed} tables is passed to Gale-Shapley deferred acceptance \citep{gale1962}, with shippers proposing: a shipper's preference over carriers ranks by closing rate ascending (cheapest first, ties broken by fewer turns to close and then a seed-derived pseudo-random order), and a carrier's preference over shippers ranks by closing rate descending (priciest first, same tie-break); impasse tables are excluded from both preference lists entirely, so an architecture that never closes a table at all is never a candidate for a match on that table.
The output is a stable matching: no shipper-carrier pair that is not matched to each other would both prefer each other over their assigned (or absent) match.
A matched carrier's \emph{margin} is $\mathrm{rate}-\bar\mu$ (the realized price minus the route's true reference-price mean); an unmatched carrier scores a margin of $0$ for that negotiation.
A matched shipper's \emph{profit} is $p_{\text{sale}}-(\mathrm{rate}+c_{\text{goods}})$; if that quantity would be negative the shipper declines the match (modeling a no-ship option) and scores a profit of $0$, as does any shipper left unmatched by Gale-Shapley.
Every negotiation thus yields exactly one margin observation per carrier and one profit observation per shipper, regardless of how many tables that agent was simultaneously negotiating.

\subsection{Nine negotiation architectures}
\label{sec:architectures}

\begin{figure}[H]
\centering
\begin{tikzpicture}[
  fam/.style={draw, rectangle, rounded corners, align=center, text width=2.5cm, minimum width=2.7cm, minimum height=1.6cm, font=\small, fill=blue!6},
  root/.style={draw, rectangle, rounded corners, align=center, text width=5cm, minimum width=5.2cm, minimum height=0.9cm, font=\small\bfseries},
  arr/.style={-{Latex[length=2mm]}},
  node distance=6mm and 4mm
]
\node[root] (root) {Nine negotiation architectures};
\node[fam, below=1.2cm of root, xshift=-8.4cm] (rule) {Rule-based\\[2pt] R, RV};
\node[fam, right=of rule] (search) {Search\\[2pt] IS};
\node[fam, right=of search] (opt) {Optimization\\[2pt] OPT, OPT2};
\node[fam, right=of opt] (ml) {Imitation\\ learning\\[2pt] ML, ML2};
\node[fam, right=of ml] (rl) {Reinforcement\\ learning\\[2pt] RL, RL2};
\draw[arr] (root.south) -- ++(0,-0.3cm) -| (rule.north);
\draw[arr] (root.south) -- ++(0,-0.3cm) -| (search.north);
\draw[arr] (root.south) -- ++(0,-0.3cm) -| (opt.north);
\draw[arr] (root.south) -- ++(0,-0.3cm) -| (ml.north);
\draw[arr] (root.south) -- ++(0,-0.3cm) -| (rl.north);
\end{tikzpicture}
\caption{The nine negotiation architectures, grouped by family. Every architecture is built in a mirrored carrier/shipper pair from the same underlying logic, with the anchors of Eqs.~\eqref{eq:carrier-band}-\eqref{eq:shipper-band} swapped per role.}
\label{fig:taxonomy}
\end{figure}
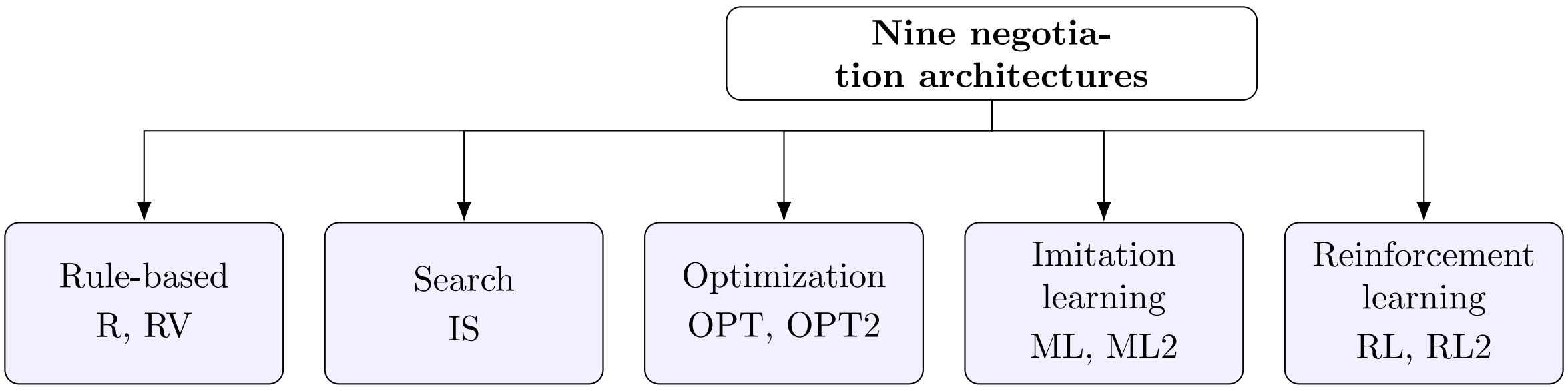

Figure~\ref{fig:taxonomy} groups the nine architectures described in this section by family.

All nine architectures share one decision interface, \texttt{next\_move(state) -> Offer | Accept | Reject}, are evaluated frozen (no online parameter updates during the tournament), and are built in mirrored pairs for the shipper and carrier role from the same underlying logic with anchors swapped per Eqs.~\eqref{eq:carrier-band}-\eqref{eq:shipper-band}; the descriptions below are given for the carrier side and apply symmetrically to the shipper side unless noted.
Every architecture shares one occupancy heuristic,
\begin{equation}
\label{eq:occupancy}
p_{\text{occ}} = \frac{1}{|\mathcal{P}|}\sum_{j\in\mathcal{P}} \mathbb{1}\!\left[\mathrm{Ready}(j)\right],
\end{equation}
over the agent's portfolio $\mathcal{P}$ of other live tables in the same negotiation, where $\mathrm{Ready}(j)$ denotes that table $j$ is already closed, or its current counterpart offer already clears the agent's own tactical threshold on that table, and one \emph{negotiation distance} that normalizes any price $p$ between a worst anchor $w$ and a best anchor $b$,
\begin{equation}
\label{eq:valuebar}
\bar v(p; w, b) = \mathrm{clip}_{[0,1]}\!\left(\frac{p-w}{b-w}\right),
\end{equation}
with $(w,b)=(RP_t,\mathrm{Tgt}_t)$ on the carrier side and $(w,b)=(RP,\mathrm{Tgt})$ on the shipper side.
The negotiation distance measures how close a price leaves the agent to the deal it wants: $\bar v=0$ is the worst deal it may legally accept, $\bar v=1$ its ambitious target.
Because it is expressed relative to each table's own band rather than in dollars, it puts routes whose reference prices differ by a factor of almost four (Eq.~\eqref{eq:route}) and the two roles, whose bands point in opposite directions, on one common 0-1 scale.
Eight of the nine architectures (all but R) are built on it, each in a different way: RV and IS use it to value closing now against continuing (Eq.~\eqref{eq:value}); OPT and OPT2 maximize an expected negotiation distance over candidate prices (Eq.~\eqref{eq:opt-accept}); ML and ML2 receive their input features and predict their offers in it; and RL and RL2 act in it and are rewarded in it (Section~\ref{sec:architectures-params}).

\subsubsection{Rule-based: R and RV}

\textbf{R} commits to no search and no learning.
Its acceptance rule compares the counterpart's current offer against a time-varying threshold
\begin{equation}
\label{eq:r-threshold}
\begin{aligned}
RP_t^{\text{eff}} &= RP_t + f\left(\mathrm{Tgt}_t - RP_t\right), \\
\theta(\tau) &= RP_t^{\text{eff}} + \left(\mathrm{Tgt}_t - RP_t^{\text{eff}}\right)\left(1-\frac{\tau}{T}\right)^{\gamma},
\end{aligned}
\end{equation}
where $f$ is a tactical-floor cushion fraction, $T$ is the number of turns available to that side, and $\gamma>1$ curves the threshold so it falls slowly at first and quickly as turns run out; the counterpart's offer is accepted once it clears $\theta(\tau)$, or unconditionally on the agent's last turn provided it clears the hard floor $RP_t$.
Absent an acceptable offer, R counters with a fixed opening anchor on the first turn and a geometrically decaying concession step thereafter,
\begin{equation}
\label{eq:r-step}
\begin{aligned}
x_1 &= \mathrm{Tgt}_t - \delta_0\sigma, \\
x_k &= x_{k-1} - \delta_1\sigma\,\rho^{\,k-2} \quad (k\ge2),
\end{aligned}
\end{equation}
clamped into $[RP_t,\mathrm{Tgt}_t]$ and into non-increasing order relative to its own last offer.

\textbf{RV} reuses R's band, threshold, and concession schedule unchanged, but replaces the time-pressure-only continue/accept comparison with the shared value function (Eq.~\eqref{eq:value}) whenever the counterpart's current offer already clears the hard floor but not the threshold $\theta(\tau)$, and scales its own concession step by an urgency factor $1+(1-p_{\text{occ}})\cdot0.5$ derived from Eq.~\eqref{eq:occupancy}, so a carrier whose other tables look unpromising concedes up to 50\% faster.
The shared value function, evaluated once for a terminal close and once for continuing at the agent's current position, is
\begin{equation}
\label{eq:value}
\begin{aligned}
V(\text{terminal}, p) &= \bar v(p; RP_t, \mathrm{Tgt}_t), \\
V(\text{continue}) &= \underbrace{\tfrac12\!\left(\bar v(\text{offer on table}) + \bar v(\text{my last offer})\right)}_{h} \\
&\quad\cdot\ \underbrace{\left(\frac{T-\tau}{T}\right)^{1/2}}_{\text{time pressure}} \,\cdot\, \underbrace{\left[(1-w_{\text{occ}}) + w_{\text{occ}}\,p_{\text{occ}}\right]}_{\text{occupancy adjustment}},
\end{aligned}
\end{equation}
with $w_{\text{occ}}\in[0,1]$ a per-engine occupancy weight; RV accepts when $V(\text{terminal})\ge V(\text{continue})$ and counters otherwise.

\subsubsection{Search-based: IS}

\textbf{IS} adapts the particle-based opponent determinization of information-set Monte Carlo tree search \citep{cowling2012} to a single bilateral decision point rather than a branching multi-ply tree, so it never performs the tree search proper and does not grow a search tree at all; a full particle-and-tree budget of roughly $10^3$ simulations per decision is impractical at the scale of this tournament.
The counterpart's hidden reservation price is determinized by sampling $M$ particles from the only information actually observable -- the counterpart's own last offer, a hard lower bound since a rule-based counterpart never offers past its own reservation price --
\begin{equation}
\label{eq:is-particles}
\begin{aligned}
\theta^{(1)},\dots,\theta^{(M)} &\sim \mathrm{Unif}\!\left(\ell,\ \ell + b\,\sigma\right), \\
\ell &= \text{counterpart's last offer (or } RP_t \text{ if none yet)},
\end{aligned}
\end{equation}
with $b$ a particle-band width constant; a set of $K$ candidate counter-offers evenly spaced between the agent's current ceiling and $RP_t$ is then scored by expected utility aggregated across every particle,
\begin{equation}
\label{eq:is-eu}
\begin{aligned}
\widehat{P}_{\text{accept}}(c) &= \frac{1}{M}\sum_{m=1}^{M}\mathbb{1}\!\left[c\le\theta^{(m)}\right], \\
\widehat{EU}(c) &= \widehat{P}_{\text{accept}}(c)\cdot V(\text{terminal},c) + \left(1-\widehat{P}_{\text{accept}}(c)\right)\cdot V(\text{continue at } c),
\end{aligned}
\end{equation}
and the candidate $c^\star=\arg\max_c\widehat{EU}(c)$ is offered, or the counterpart's current offer is accepted instead whenever its own terminal value already meets or beats $\widehat{EU}(c^\star)$.

\subsubsection{Continuous optimization: OPT and OPT2}

\textbf{OPT} treats the next offer as a real-valued variable $x\in[RP_t,\text{ceiling}]$ and climbs an expected-utility surface directly rather than ranking a small candidate set.
A belief about the counterpart's current acceptance threshold $\hat\theta$ (the counterpart's last offer plus a fixed cushion) feeds a differentiable acceptance-probability model,
\begin{equation}
\label{eq:opt-accept}
\begin{aligned}
P_{\text{accept}}(x) &= \mathrm{sigmoid}\!\left(-\beta\,\frac{x-\hat\theta}{\sigma}\right), \\
EU(x) &= P_{\text{accept}}(x)\,\bar v(x;RP_t,\mathrm{Tgt}_t) + \left(1-P_{\text{accept}}(x)\right)V(\text{continue at } x),
\end{aligned}
\end{equation}
with $\beta$ a steepness constant.
Both terms of $EU$ are negotiation distances (Eq.~\eqref{eq:valuebar}): the first directly, the second through the continuation value of Eq.~\eqref{eq:value}, which is itself built from the distances of the offer on the table and of the agent's own last offer.
$EU(x)$ is therefore the negotiation distance the agent expects to end at if it offers $x$, and the optimum
\begin{equation}
\label{eq:bisect}
x^\star = \arg\max_{x\in[RP_t,\,\text{ceiling}]} EU(x)
\end{equation}
is located by bisecting on the \emph{sign} of a central-difference gradient at the window's midpoint $m$,
\begin{equation}
\label{eq:bisect-grad}
\begin{aligned}
g(m) &= \frac{EU(m+h)-EU(m-h)}{2h}, \\
h &= \max\!\left(0.01\,(\text{hi}-\text{lo}),\,10^{-9}\right),
\end{aligned}
\end{equation}
narrowing the window to $[m,\text{hi}]$ if $g(m)>0$ and to $[\text{lo},m]$ otherwise, for a fixed number of iterations, rather than stepping by the gradient's magnitude, since $EU$ lives on the 0-1 negotiation-distance scale, unrelated to the dollar scale being searched.
$x^\star$ is thus the price whose expected negotiation distance is highest, and OPT's accept rule compares like with like: it accepts the counterpart's current offer whenever that offer's own negotiation distance already meets or beats $EU$ at the offer OPT would otherwise make.
A final offer is then emitted a small, reproducibly-seeded distance below $x^\star$ in the concession direction only, so that a deterministic optimizer is not trivially fingerprintable by a counterpart observing many negotiations.

\textbf{OPT2} keeps OPT's optimizer and acceptance model but replaces its single-point belief with a line
$\text{offer}_k \sim a + b\,k$
fitted to the counterpart's entire observed offer sequence by explicit gradient descent on squared error over the $(a,b)$ parameters, extrapolated one step beyond the last observed offer and then cushioned exactly as OPT's point estimate is; it also replaces OPT's fixed-magnitude jitter with a performance-neutral randomization envelope, widening a window downward from $x^\star$ for as long as $EU$ stays within a fixed fractional tolerance $\epsilon_{\text{tol}}$ of $EU(x^\star)$,
\begin{equation}
\label{eq:opt2-envelope}
\begin{aligned}
a &= \inf\{x\le x^\star : EU(x^\star)-EU(x)\le\epsilon_{\text{tol}}\,|EU(x^\star)|\}, \\
\text{window} &= \left[\,a,\ x^\star\,\right],
\end{aligned}
\end{equation}
and emitting an offer sampled uniformly from that window with a seed derived from the negotiation, so that the resulting randomization never costs expected utility by more than $\epsilon_{\text{tol}}$ while still varying the realized offer.

\subsubsection{Imitation learning: ML and ML2}

\textbf{ML} works entirely in negotiation distance (Eq.~\eqref{eq:valuebar}): every price-like input feature (offers on the table, the agent's own last offer) is expressed as a distance, and the model predicts, from the current table state, the negotiation distance $p\in[0,1]$ at which to make its next offer, via a gradient-boosted regressor trained on OPT2's self-play offers \citep{pedregosa2011} (Section~\ref{sec:architectures-params} gives training details), maps it to a price $x = RP_t + p\,(\mathrm{Tgt}_t-RP_t)$, and clamps $x$ into the same legal band and monotonicity constraint every other engine respects; the accept/continue decision is deliberately not left to the model and instead reuses the shared value-function comparison of Eq.~\eqref{eq:value}, compared against the agent's own already-stated last offer rather than its freshly predicted $x$, since comparing against a hypothetical, not-yet-offered $x$ was found to trigger accepting far too early.

\textbf{ML2} extends ML with a second, independent classifier that also predicts whether to accept the counterpart's current offer, $\widehat{P}_{\text{accept}}(\text{state})$, trained separately from the offer regressor so that neither model's output is fed into the other as an input; the offer regressor and the hard last-turn floor behave exactly as in ML, and ML2 accepts an otherwise-legal counterpart offer whenever $\widehat{P}_{\text{accept}}(\text{state})\ge0.5$.

\subsubsection{Reinforcement learning: RL and RL2}

\textbf{RL} is a policy trained with proximal policy optimization \citep{schulman2017,raffin2021} against a single-agent negotiation environment in which the policy's reward depended only on closing a legal deal in that isolated environment, with no rival carriers present during training.
At inference the policy outputs a single aspiration $p\in[0,1]$, a negotiation distance (Eq.~\eqref{eq:valuebar}) mapped to a price exactly as in ML/ML2, and the agent accepts the counterpart's current offer outright whenever that offer already clears both the hard floor and the policy's own aspiration, rather than deferring to a hand-derived value-function comparison.

\textbf{RL2} is trained with the same action mechanism and the same algorithm, but its training reward depended on actually winning the Gale-Shapley matching stage (Section~\ref{sec:matching}) against rival carriers, not merely on closing a deal in isolation.
Tracing a deployed round, however, surfaced a confirmed, regression-tested bug in RL2's inference path: its training environment always held the policy's own portfolio-occupancy feature fixed at a neutral default, since the learning agent's own table never had real siblings during training, while a deployed RL2 genuinely negotiates several tables at once and would otherwise present that feature at whatever its real, non-neutral portfolio state is; RL2's inference code pins the feature back to its neutral training value before every decision, restoring the input distribution the policy was actually trained on rather than feeding it an input it never saw in training.
We deliberately evaluate RL2 as actually deployed, with this feature pinned, rather than silently repairing the training/deployment mismatch itself, since the gap between what an architecture could in principle do and what the tournament actually measures is itself one of this paper's results (Section~\ref{sec:results-rl2gap}).

\subsection{Architecture parameters and training}
\label{sec:architectures-params}

Table~\ref{tab:hyperparams} lists each architecture's governing hyperparameters, held fixed across every market condition and role (Section~\ref{sec:discussion} returns to the consequences of not tuning them further).
\textbf{ML} and \textbf{ML2}'s offer regressors are gradient-boosted regression trees trained to imitate \textbf{OPT2} specifically -- not a from-scratch policy or a hand-authored oracle -- on several hundred self-play negotiations of OPT2 against every other hand-designed architecture, evaluated by held-out mean absolute error against OPT2's own band position; \textbf{ML2}'s accept classifier is a second, independently trained gradient-boosted classifier on the same feature vector, fit only on turns where accepting was legal, and never fed the offer regressor's output or vice versa.
\textbf{RL} and \textbf{RL2} are both trained with proximal policy optimization \citep{schulman2017,raffin2021} (a two-layer, 64-unit multilayer-perceptron policy, discount factor $\gamma=0.99$, for $300{,}000$ environment steps), differing only in the environment and reward that training optimizes against: RL's terminal reward is the negotiation distance of the closing price, $\bar v(\text{closing price}; RP_t, \mathrm{Tgt}_t)$ (Eq.~\eqref{eq:valuebar}), whenever its one training opponent for that episode (resampled each episode from \{R, RV, IS, OPT, OPT2\}) accepts a legal deal in an isolated, single-table environment, and zero otherwise; RL2 is trained inside a small competitive sub-round of one to two same-side rivals (sampled from the same five-architecture pool) against one shared counterpart, cleared by the same Gale-Shapley mechanism as the full tournament (Section~\ref{sec:matching}), with reward $-0.5$ on impasse, $-0.15$ if its own table closes but the match assigns the counterpart to a rival instead, and $\bar v(\text{closing price}; RP_t, \mathrm{Tgt}_t)$ conditional on winning the match.

\begin{table}[H]
\centering
\caption{Governing hyperparameters for all nine architectures, held fixed across every market condition and role.}
\label{tab:hyperparams}
\small
\begin{tabular}{@{}l p{0.82\linewidth}@{}}
\toprule
Architecture & Key parameters \\
\midrule
R & Opening offset $0.15\sigma$ (carrier) / band width $0.5\sigma$ (shipper); concession decay $\rho=0.5$; threshold curvature $\gamma=2.0$ \\
RV & As R, plus the occupancy-aware urgency factor derived from Eq.~\eqref{eq:occupancy} (up to 50\% faster concession when the agent's own match prospects look poor) \\
IS & 24 belief particles; 7 candidate offers evaluated per turn; particle-sampling band $1.5\sigma$ wide; deterministic per-decision seeding (no run-to-run variance at a fixed seed) \\
OPT & Continuous bisection search over expected utility (24 steps); acceptance-probability sigmoid slope $4.0$; belief cushion $0.5\sigma$ \\
OPT2 & Extends OPT with an online gradient-fitted linear opponent-offer-trend model (60 gradient steps, learning rate $0.15$) and an anti-fingerprinting envelope search (200-iteration cap, $5\%$ expected-utility tolerance) \\
ML & Gradient-boosted regression tree (max depth 4, 200 boosting iterations, learning rate $0.08$), trained to imitate OPT2's self-play offers \\
ML2 & Same offer regressor as ML, plus an independently trained gradient-boosted accept classifier (same hyperparameters), decision threshold $0.5$ \\
RL & PPO, $2\times64$ MLP policy, $\gamma=0.99$, $300{,}000$ training steps, one isolated opponent per episode resampled from \{R, RV, IS, OPT, OPT2\} \\
RL2 & Same PPO configuration as RL, trained inside a Gale-Shapley-cleared competitive sub-round against one to two same-side rivals from the same five-architecture pool \\
\bottomrule
\end{tabular}
\end{table}

\subsection{Experimental design}
\label{sec:design}

\begin{figure}[H]
\centering
\begin{tikzpicture}[
  arch/.style={draw, circle, minimum size=0.9cm, font=\scriptsize, align=center},
  excl/.style={draw, circle, minimum size=0.9cm, font=\scriptsize, align=center, fill=gray!25, draw=gray, text=gray},
  node distance=2mm
]
\node[arch] (r) {R};
\node[arch, right=of r] (rv) {RV};
\node[arch, right=of rv] (is) {IS};
\node[excl, right=of is] (opt) {OPT};
\node[arch, right=of opt] (opt2) {OPT2};
\node[arch, right=of opt2] (ml) {ML};
\node[excl, right=of ml] (ml2) {ML2};
\node[arch, right=of ml2] (rl) {RL};
\node[arch, right=of rl] (rl2) {RL2};
\draw[-{Latex[length=1.8mm]}] (opt.south) to[bend right=40] node[below,font=\scriptsize]{excluded this sub-round} (ml2.south);
\end{tikzpicture}
\caption{One of the $\binom{9}{2}=36$ exclusion pairs used to build a scarce condition: the nine architectures on the scarce side, with two (shaded; OPT and ML2 shown here purely for illustration) withheld for that sub-round's $N_{\text{pair}}=12$ negotiations while the other seven compete and all nine architectures remain present on the abundant side. The excluded pair cycles through all 36 combinations in turn, so every architecture is withheld from exactly 8 of the 36 sub-rounds and present in the remaining 28, per Eq.~\eqref{eq:neff}.}
\label{fig:design}
\end{figure}
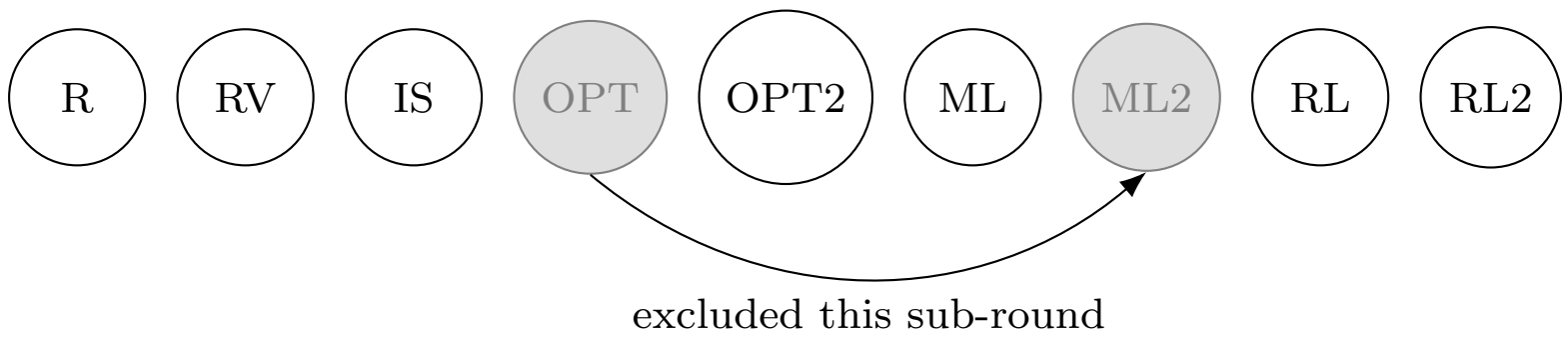

Figure~\ref{fig:design} illustrates the exclusion-pair rotation scheme used to build each scarce condition.

Every architecture appears at most once per side per negotiation, so that no result reflects one architecture competing against a copy of itself rather than against the other eight.
The \emph{balanced} condition runs $N=400$ negotiations with all nine architectures present on both sides.
Each \emph{scarce} condition reduces one side's roster from nine architectures to seven, holding the other side at all nine; which two architectures are excluded is rotated systematically through every one of the $\binom{9}{2}=36$ possible exclusion pairs, running $N_{\text{pair}}=12$ negotiations per excluded pair, for $36\times12=432$ negotiations total per scarce condition.
Because each architecture is excluded only in the 8 pairs that name it (one pairing with each of the other eight architectures), it is present in $36-8=28$ of the 36 sub-rounds, contributing an effective sample size on the scarce side of
\begin{equation}
\label{eq:neff}
N_{\text{eff,scarce}} = N_{\text{pair}}\times 28 = 336,
\end{equation}
while the abundant side of the same scarce condition, present in every one of the 36 sub-rounds, contributes the full $N_{\text{pair}}\times36=432$ observations per architecture; both figures are counted directly from the resulting data rather than assumed, and matched the values above exactly.
We refer to the condition in which carriers are reduced to seven architectures as \texttt{carrier\_scarce} (shippers compete for scarce carrier capacity) and the mirror condition as \texttt{shipper\_scarce} (carriers compete for scarce shipper demand).

\subsection{Computational setup}
\label{sec:hardware}

All experiments -- the nine-architecture tournament, the training of ML/ML2 and RL/RL2, and the language-model comparison of Section~\ref{sec:results-llm} -- ran on a single workstation: an Intel Core i9-14900F CPU (24 cores, 32 threads), 128\,GB of RAM, and one NVIDIA GeForce RTX 4090 GPU with 24\,GB of memory, under Windows 11 Pro.
RL2 was trained on the CPU; RL used the training library's default device selection, which picks the GPU when one is available \citep{raffin2021}.
Language models were served locally through Ollama with each model tag's default quantization and no other request options; the exact Ollama server version in use during data collection was not logged (the installed client is version 0.34.2).
This matters for the language-model results: at that quantization each of the three 70B-class models (\texttt{llama3.3:70b}, \texttt{qwen2.5:72b}, \texttt{deepseek-r1:70b}) needs more memory than the GPU's 24\,GB, so part of each model ran from CPU memory, which slows inference and makes the latencies reported in Section~\ref{sec:results-llm} specific to this hardware.

\subsection{Statistical analysis}
\label{sec:stats}

Every architecture's outcome under a given market condition and role is summarized by a percentile bootstrap confidence interval rather than a normal-approximation interval, since margin and profit outcomes are not symmetrically distributed: given $n$ observations $y_1,\dots,y_n$, we resample $B=5000$ bootstrap samples of size $n$ with replacement, compute the sample mean $\bar y^{(b)}$ of each resample, let $q_\alpha$ denote the $\alpha$-quantile of the resulting set $\{\bar y^{(b)}\}_{b=1}^{B}$, and report
\begin{equation}
\label{eq:bootstrap}
\widehat{\mathrm{CI}}_{95\%} = \left[\,q_{0.025},\ q_{0.975}\,\right].
\end{equation}
For every pair of architectures present in the same market condition and role, we restrict to the subset of negotiation-days on which \emph{both} architectures were actually present (every day, in the balanced condition; the subset of sub-rounds excluding neither architecture, in a scarce condition), form the paired per-day difference $d_i = y_i^{(A)}-y_i^{(B)}$, and test $H_0:\mathrm{median}(d)=0$ with a two-sided Wilcoxon signed-rank test, a non-parametric choice appropriate for the same skewed outcome distributions the bootstrap CI accounts for.
Because every market condition and role admits $\binom{9}{2}=36$ such pairwise comparisons answering the same family of questions (``who beats whom under this market condition''), we correct the 36 raw $p$-values within each (market condition, role) family separately with the Holm-Bonferroni step-down procedure at $\alpha=0.05$: sorting the family's $p$-values $p_{(1)}\le\dots\le p_{(36)}$, a hypothesis at rank $i$ is rejected only if $p_{(j)}\le\alpha/(36-j+1)$ for every $j\le i$, which controls the family-wise error rate without assuming independence between the 36 tests.

\section{Results}
\label{sec:results}

\subsection{Descriptive ranking by market condition}
\label{sec:results-ranking}

\begin{figure}[H]
\centering
\includegraphics[width=\textwidth]{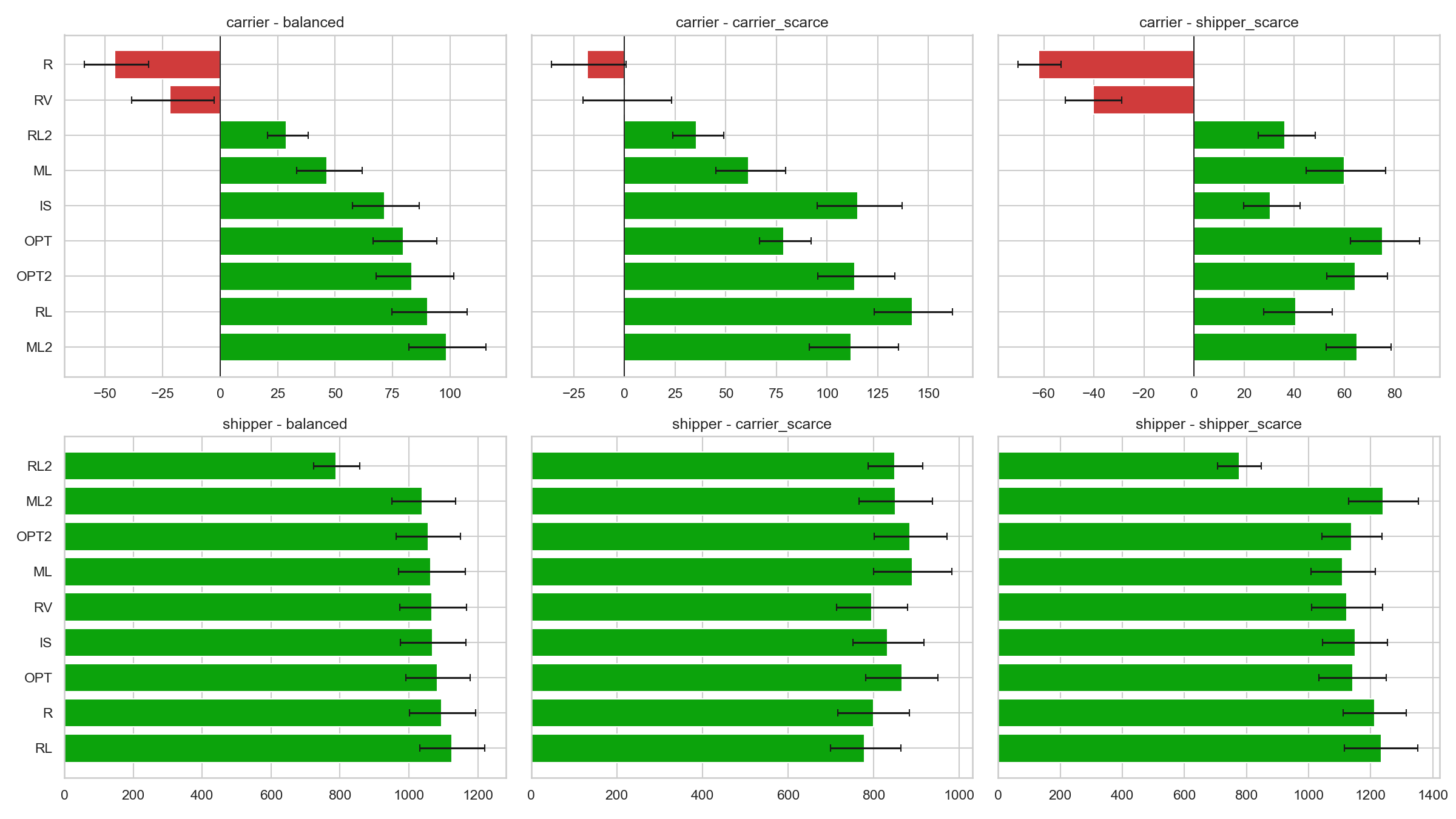}
\caption{Bootstrap mean outcome (margin for carriers, profit for shippers) by architecture, for all nine architectures under all three market conditions. Error bars are 95\% percentile bootstrap confidence intervals.}
\label{fig:ranking}
\end{figure}

Table~\ref{tab:carrier} and Table~\ref{tab:shipper} report each architecture's match rate and bootstrap mean outcome (margin for carriers, profit for shippers) in every market condition, both in USD and as a percentage that puts the dollar figures in scale: ML2's balanced-condition mean margin of 98.5\,USD, for example, is 6.6\% of the matched route's reference price, and a shipper's typical profit of roughly 1,100\,USD is about 20\% of its sale price; Figure~\ref{fig:ranking} plots the same rankings with their confidence intervals.
In the balanced carrier market, the top of the ranking mixes three different families rather than one: by point estimate the imitation-learned ML2 is followed by the reinforcement-learned RL, then the continuous-optimization OPT2 and OPT, but none of these four differs significantly from any other after correction (Section~\ref{sec:results-pairwise}), so they form a statistically tied top group rather than a strict order; the search-based IS sits in the upper-middle, and the two rule-based architectures R and RV occupy the bottom two positions, both with a mean margin confidence interval that excludes zero on the losing side, R by the widest margin of any architecture.
In the balanced shipper market the ordering is far more compressed among the top eight architectures (RL through ML2, all within roughly 85 profit units of each other) with one striking exception: RL2 is isolated at the bottom of the shipper ranking by a margin of roughly 250-340 profit units in the balanced and \texttt{shipper\_scarce} conditions, a gap large enough to be visible in Table~\ref{tab:shipper} without any statistical test; under \texttt{carrier\_scarce}, however, RL2 moves into the middle of the ranking (849.8, fifth of nine) and it is instead RL that falls to the bottom (779.2), a reordering discussed further in Section~\ref{sec:results-stability}.

\begin{table}[H]
\centering
\caption{Carrier-side match rate and bootstrap mean margin in USD (95\% CI) by architecture and market condition. $N$ is the effective sample size (Eq.~\eqref{eq:neff} for the scarce side of a scarce condition). ``\% of ref.'' is the mean, over negotiation-days, of the margin as a percentage of the matched route's reference price $\bar\mu$ (an unmatched day counts as 0\%); because expensive routes weigh more in dollars than in percent, the two columns can rank close architectures differently (e.g.\ ML2 and RL in the balanced condition). 77 of the tournament's 8,256 matched carrier-days (0.9\%) could not be paired with their route from the stored logs and are left out of this column only.}
\label{tab:carrier}
\small
\resizebox{\textwidth}{!}{%
\begin{tabular}{lrrrrrrrrrrrr}
\toprule
 & \multicolumn{4}{c}{balanced ($N=400$)} & \multicolumn{4}{c}{carrier\_scarce ($N=336$)} & \multicolumn{4}{c}{shipper\_scarce ($N=432$)} \\
\cmidrule(lr){2-5}\cmidrule(lr){6-9}\cmidrule(lr){10-13}
Arch. & match & mean margin (USD) & \% of ref. & 95\% CI & match & mean margin (USD) & \% of ref. & 95\% CI & match & mean margin (USD) & \% of ref. & 95\% CI \\
\midrule
ML2  & 0.79 & 98.5  & 6.6 & [82.2, 115.6]  & 0.84 & 111.9 & 8.0 & [91.2, 135.1]  & 0.62 & 65.1 & 4.5 & [52.9, 78.7] \\
RL   & 0.62 & 90.4  & 6.9 & [74.6, 107.5]  & 0.84 & 142.0 & 10.6 & [123.3, 162.0] & \textbf{0.16} & 40.8 & 3.2 & [27.8, 55.2] \\
OPT2 & 0.80 & 83.5  & 5.9 & [67.7, 101.7]  & 0.91 & 113.8 & 7.3 & [95.4, 133.4]  & 0.69 & 64.4 & 4.4 & [53.0, 77.3] \\
OPT  & 0.79 & 79.8  & 5.3 & [66.5, 94.3]   & 0.87 & 78.7  & 4.9 & [66.8, 92.1]   & 0.76 & 75.4 & 5.1 & [62.5, 90.1] \\
IS   & 0.69 & 71.4  & 5.0 & [57.6, 86.5]   & 0.82 & 115.2 & 8.1 & [95.0, 136.9]  & \textbf{0.23} & 30.6 & 2.6 & [19.8, 42.4] \\
ML   & 0.89 & 46.4  & 3.2 & [33.4, 61.9]   & 0.94 & 61.4  & 3.9 & [45.2, 79.7]   & 0.86 & 59.9 & 4.4 & [44.7, 76.5] \\
RL2  & 0.92 & 28.7  & 1.6 & [20.6, 38.2]   & 0.93 & 35.5  & 2.2 & [23.7, 48.9]   & 0.92 & 36.4 & 2.1 & [25.7, 48.5] \\
RV   & 0.92 & $-$21.8 & 0.0 & [$-$38.5, $-$2.6] & 0.93 & 0.2 & 1.3 & [$-$20.4, 23.2] & 0.89 & $-$40.4 & $-$1.3 & [$-$51.4, $-$28.8] \\
R    & 0.93 & $-$45.8 & $-$1.6 & [$-$59.0, $-$31.0] & 0.93 & $-$18.5 & 0.2 & [$-$36.0, 0.8] & 0.94 & $-$62.1 & $-$2.8 & [$-$70.2, $-$53.0] \\
\bottomrule
\end{tabular}%
}
\end{table}

\begin{table}[H]
\centering
\caption{Shipper-side match rate and bootstrap mean profit in USD (95\% CI) by architecture and market condition. ``\% of sale'' is the mean, over negotiation-days, of the profit as a percentage of that day's sale price $p_{\text{sale}}$ (an unmatched or declined day counts as 0\%).}
\label{tab:shipper}
\small
\resizebox{\textwidth}{!}{%
\begin{tabular}{lrrrrrrrrrrrr}
\toprule
 & \multicolumn{4}{c}{balanced ($N=400$)} & \multicolumn{4}{c}{carrier\_scarce ($N=432$)} & \multicolumn{4}{c}{shipper\_scarce ($N=336$)} \\
\cmidrule(lr){2-5}\cmidrule(lr){6-9}\cmidrule(lr){10-13}
Arch. & match & mean profit (USD) & \% of sale & 95\% CI & match & mean profit (USD) & \% of sale & 95\% CI & match & mean profit (USD) & \% of sale & 95\% CI \\
\midrule
RL   & 0.82 & 1124.8 & 20.8 & [1032.4, 1220.5] & 0.63 & 779.2 & 14.2 & [699.9, 863.6]  & 0.85 & 1233.7 & 21.7 & [1115.5, 1352.1] \\
R    & 0.82 & 1094.4 & 20.1 & [1002.5, 1193.9] & 0.66 & 800.5 & 14.8 & [716.2, 883.9]  & 0.89 & 1212.9 & 21.9 & [1111.6, 1315.5] \\
OPT  & 0.84 & 1082.5 & 20.0 & [990.3, 1177.8]  & 0.69 & 867.1 & 15.9 & [781.3, 951.3]  & 0.87 & 1143.5 & 20.0 & [1033.2, 1250.3] \\
IS   & 0.79 & 1068.9 & 19.6 & [976.0, 1165.3]  & 0.70 & 833.1 & 15.3 & [752.5, 918.0]  & 0.85 & 1149.7 & 20.8 & [1044.5, 1253.1] \\
RV   & 0.80 & 1066.9 & 19.5 & [973.1, 1167.3]  & 0.66 & 795.4 & 14.6 & [713.8, 879.7]  & 0.86 & 1122.9 & 19.9 & [1010.5, 1237.6] \\
ML   & 0.83 & 1063.9 & 19.6 & [969.6, 1163.2]  & 0.71 & 891.6 & 16.1 & [800.7, 982.9]  & 0.87 & 1109.8 & 20.1 & [1008.0, 1215.0] \\
OPT2 & 0.82 & 1055.7 & 19.3 & [963.4, 1149.8]  & 0.69 & 885.0 & 16.2 & [801.3, 971.7]  & 0.87 & 1139.6 & 21.0 & [1042.4, 1236.2] \\
ML2  & 0.83 & 1038.8 & 19.1 & [950.2, 1135.2]  & 0.69 & 852.0 & 15.6 & [766.7, 938.1]  & 0.88 & 1240.8 & 22.0 & [1128.4, 1354.3] \\
RL2  & 0.81 & \textbf{788.2} & 14.5 & [722.8, 857.0] & 0.81 & \textbf{849.8} & 15.6 & [787.8, 914.6]  & 0.84 & \textbf{776.8} & 14.3 & [706.7, 848.2] \\
\bottomrule
\end{tabular}%
}
\end{table}

\subsection{Pairwise significance after correction}
\label{sec:results-pairwise}

\begin{figure}[H]
\centering
\includegraphics[width=0.55\textwidth]{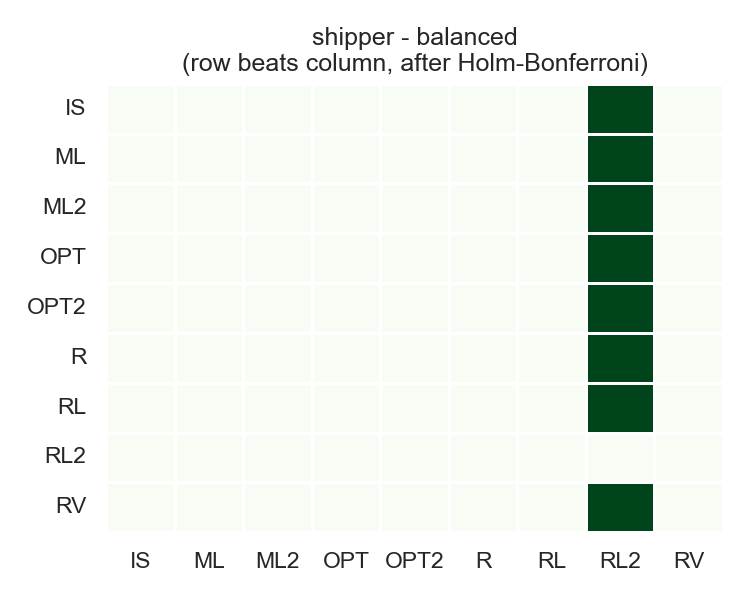}
\caption{Shipper-side win matrix, balanced market condition: a filled cell in row $i$, column $j$ means architecture $i$ significantly outperforms architecture $j$ after Holm-Bonferroni correction. RL2 wins zero of its eight comparisons.}
\label{fig:winmatrix-shipper-balanced}
\end{figure}

After Holm-Bonferroni correction within each (market condition, role) family of 36 comparisons, 27 of 36 carrier comparisons were significant in the balanced condition, 28 of 36 in \texttt{carrier\_scarce}, and 30 of 36 in \texttt{shipper\_scarce}; on the shipper side, 8 of 36 comparisons were significant in the balanced condition, \emph{zero} of 36 in \texttt{carrier\_scarce}, and 8 of 36 in \texttt{shipper\_scarce}.
The complete absence of significant shipper-side differences under \texttt{carrier\_scarce} is itself a finding: once shippers must compete for a reduced pool of carriers, no shipper architecture's negotiating strategy produces a statistically distinguishable advantage in realized profit over any other, even though the same nine architectures are clearly separated from each other in the balanced and \texttt{shipper\_scarce} conditions.
Figure~\ref{fig:winmatrix-shipper-balanced} shows the balanced-condition shipper win matrix: RL2 loses every one of its eight significant comparisons and wins none, the only architecture with a completely one-sided record in the balanced condition in either role.

\begin{figure}[H]
\centering
\begin{subfigure}[b]{0.32\textwidth}
  \centering
  \includegraphics[width=\textwidth]{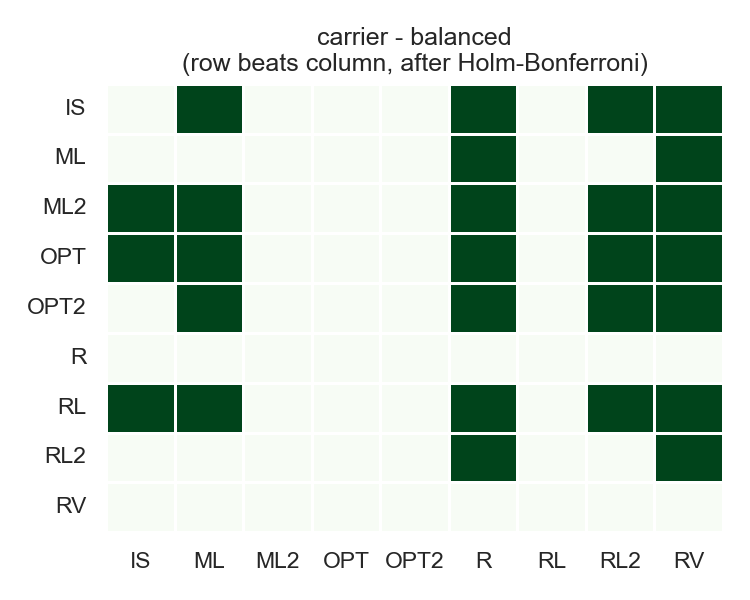}
  \caption{carrier, balanced}
\end{subfigure}
\hfill
\begin{subfigure}[b]{0.32\textwidth}
  \centering
  \includegraphics[width=\textwidth]{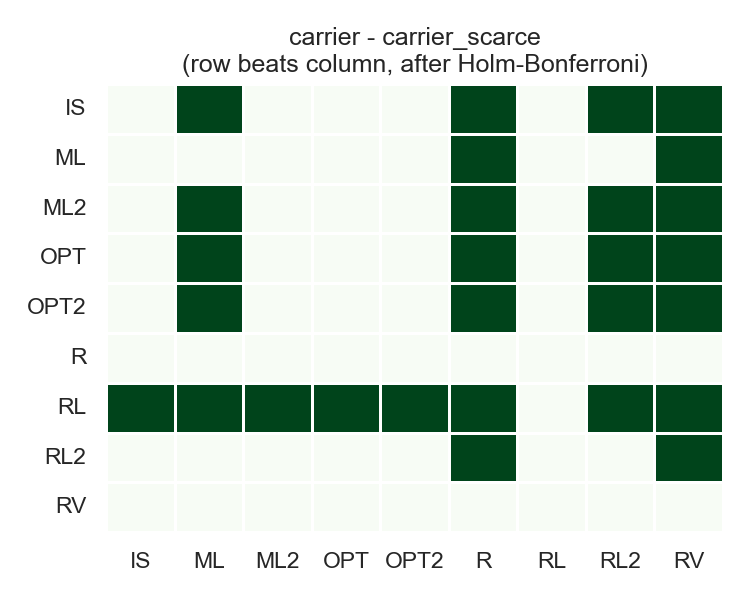}
  \caption{carrier, carrier\_scarce}
\end{subfigure}
\hfill
\begin{subfigure}[b]{0.32\textwidth}
  \centering
  \includegraphics[width=\textwidth]{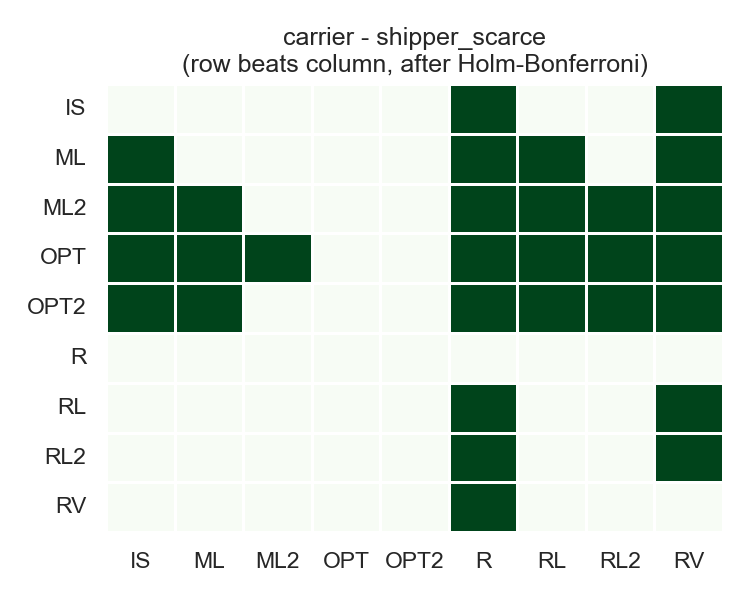}
  \caption{carrier, shipper\_scarce}
\end{subfigure}

\vspace{0.4cm}

\begin{subfigure}[b]{0.32\textwidth}
  \centering
  \includegraphics[width=\textwidth]{figures/win_matrix_shipper_balanced.png}
  \caption{shipper, balanced}
\end{subfigure}
\hfill
\begin{subfigure}[b]{0.32\textwidth}
  \centering
  \includegraphics[width=\textwidth]{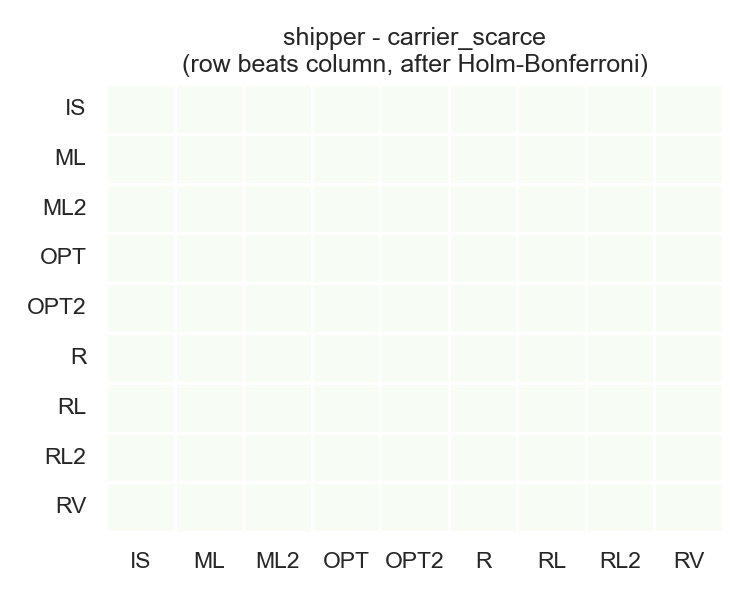}
  \caption{shipper, carrier\_scarce}
\end{subfigure}
\hfill
\begin{subfigure}[b]{0.32\textwidth}
  \centering
  \includegraphics[width=\textwidth]{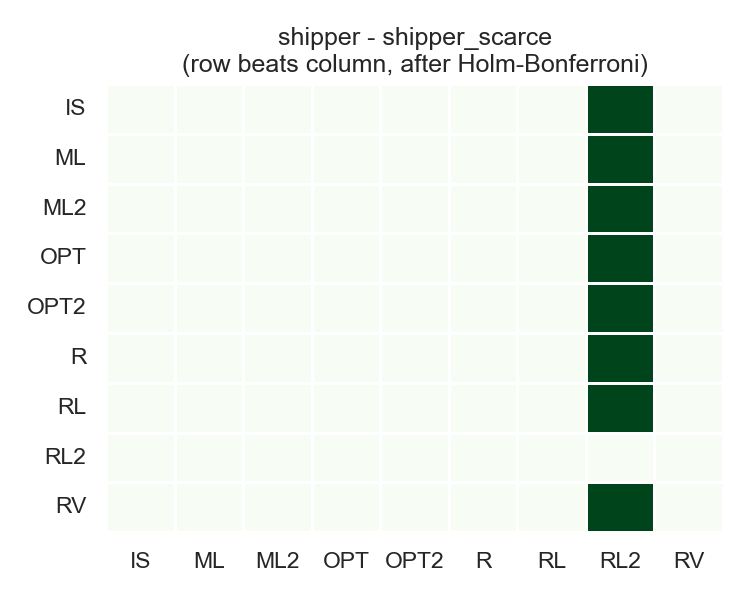}
  \caption{shipper, shipper\_scarce}
\end{subfigure}
\caption{All six pairwise win matrices (row significantly beats column, after Holm-Bonferroni correction within each panel's 36 comparisons): carrier role (top row) and shipper role (bottom row), across the three market conditions. Panel (d) is reproduced larger as Figure~\ref{fig:winmatrix-shipper-balanced} above.}
\label{fig:winmatrix-all}
\end{figure}

Figure~\ref{fig:winmatrix-all} reproduces all six win matrices (both roles, all three market conditions) for completeness; panel (f) (shipper, \texttt{shipper\_scarce}) and panel (c) (carrier, \texttt{shipper\_scarce}) are the two panels most relevant to the match-rate collapse of Section~\ref{sec:results-collapse}, since RL and IS's matching failures there mean the two architectures contribute very few observations to panel (c)'s comparisons even where they remain nominally present.

\subsection{Competitive training and match-rate robustness under scarcity}
\label{sec:results-collapse}

Table~\ref{tab:carrier} shows that two carrier architectures, RL and IS, lose the large majority of their matching opportunities specifically under \texttt{shipper\_scarce}, the one condition in which carriers must genuinely out-compete each other for a reduced pool of shipper demand: RL's match rate falls from 0.62 (balanced) and 0.84 (\texttt{carrier\_scarce}, where carriers are \emph{not} competing with each other for demand) to \textbf{0.16} under \texttt{shipper\_scarce}, and IS falls from 0.69 and 0.82 to \textbf{0.23} under the same condition.
No other carrier architecture shows a comparable drop: RL2's match rate instead stays within a narrow band (0.92, 0.93, 0.92) across all three conditions. R (0.93, 0.93, 0.94) and RV (0.92, 0.93, 0.89) are comparably stable, however, despite neither being trained with any competitive pressure at all; match-rate stability on its own therefore does not establish that RL2's stability is caused by its training against rival carriers, only that RL2 avoids the specific failure mode that collapses RL and IS.
RL was trained with proximal policy optimization against an environment whose reward depended only on closing a deal in an isolated, non-competitive setting (Section~\ref{sec:architectures}); IS's opponent model is a one-shot particle determinization of a single counterpart's reservation price, with no representation of rival carriers at all.
Neither architecture's training or search procedure ever represented the possibility of losing a deal to a rival carrier after agreeing a price with a shipper, and Table~\ref{tab:carrier} shows the direct cost of that gap exactly in the one market condition designed to expose it.

\subsection{The RL2 train/deployment gap}
\label{sec:results-rl2gap}

RL2 avoids the match-rate collapse of Section~\ref{sec:results-collapse}, but Table~\ref{tab:carrier} and Table~\ref{tab:shipper} show it ranks at or near the bottom of realized outcome in the balanced and \texttt{shipper\_scarce} conditions, despite that stable match rate; under \texttt{carrier\_scarce} it instead rises to the middle of the shipper ranking, a reordering Section~\ref{sec:results-stability} discusses in full.
On the shipper side, every one of RL2's eight pairwise comparisons in the balanced condition is significant after correction, all favoring the rival architecture, with $p_{\text{holm}}<10^{-12}$ in every case (Section~\ref{sec:results-pairwise}, Figure~\ref{fig:winmatrix-shipper-balanced}); on the carrier side, seven of RL2's eight balanced-condition comparisons are significant and all favor the rival, the sole exception (RL2 vs.\ ML, mean difference $17.7$, $p_{\text{holm}}=1.0$) not reaching significance after correction.
This outcome is consistent with the confirmed feature mismatch described in Section~\ref{sec:architectures}: because RL2's training environment always held its own portfolio-occupancy feature at a fixed neutral value, and the pinned inference-time value reproduces that same neutral value rather than differing from it, the policy never had an opportunity to learn how to moderate its aspiration in response to genuine competitive pressure; a trace of a deployed round confirmed the shipper-side policy opening at roughly 94\% of its own worst-case ceiling on the very first turn and being accepted instantly by whichever counterpart it faced, the behavior of a policy whose aspiration was shaped entirely by a training signal that never varied with real portfolio risk.
RL2 therefore demonstrates that training a policy against genuine competitive pressure can solve the specific failure mode of Section~\ref{sec:results-collapse} (match-rate collapse under scarcity) without producing a policy that is competitive on realized payoff, if a feature the policy depends on is not presented identically at training and at deployment time.

\subsection{Rank stability of margin and profit across market conditions}
\label{sec:results-stability}

Every ranking in this subsection orders architectures by realized economic outcome -- mean margin for carriers, mean profit for shippers, in USD (Tables~\ref{tab:carrier}-\ref{tab:shipper}) -- not by match rate.
The carrier-side top four under the balanced condition (ML2, RL, OPT2, OPT) remain in the top five under both scarce conditions, though their relative order shifts: RL rises to first under \texttt{carrier\_scarce} and IS rises into the top four under both scarce conditions, while OPT and ML2 fall in relative rank specifically under \texttt{shipper\_scarce} even though their mean margin changes comparatively little.
On the shipper side, the top eight architectures reorder modestly across conditions (R and RL exchange the top position; ML2 rises from eighth under balanced to first under \texttt{shipper\_scarce}), and RL2 is not uniformly the bottom performer: it is last in the balanced and \texttt{shipper\_scarce} conditions by the same large margin documented in Section~\ref{sec:results-pairwise}, but under \texttt{carrier\_scarce} it rises to fifth of nine (849.8) while RL itself drops to the bottom (779.2) -- RL's own worst result of the three conditions, and the only condition in which RL2 is not the shipper-side floor.

\subsection{Comparison against LLM-based negotiation agents}
\label{sec:results-llm}

As a supplementary extension of the same tournament methodology, we additionally embedded large-language-model-based negotiation agents (served locally via Ollama, prompted with the agent's exact hard-band numbers and its own table's full turn history, using the same legality and monotonicity enforcement as every architecture in Section~\ref{sec:architectures}) into a small market alongside three fixed same-side rivals spanning this tournament's weakest, strongest hand-designed, and strongest learned carrier architectures (R, OPT2, and ML2), against a fixed four-architecture counterpart roster (R, RV, OPT2, IS), across nine language models spanning three parameter-count tiers (tiny, small, and medium: under 3B, 4-14B, and 15-70B parameters respectively) and three model families (Llama, Qwen, and a DeepSeek-R1 distillation) at each tier, for 40 independent negotiation-days per (model, role) combination, with each call bound by a 90-second request timeout and a \texttt{max\_retries=2} budget before that day forfeited.
Two methodological notes before the results, reported transparently rather than silently absorbed.
First, an unattended multi-day run that must survive repeated crashes and restarts -- realized by resuming from whichever days a prior process already wrote -- turned up a seeding bug this comparison alone exposed: the per-(model,role) combination seed was originally derived from Python's built-in \texttt{hash()} on a string tuple, which is randomized per process by design and therefore not reproducible across the many restarts this study required; we traced it after finding genuinely divergent outcomes recorded under the same day index in two (model, role) files, fixed it with a process-independent hash, and deduplicated the affected files down to one valid realization per day before any statistic below was computed.
Same-day comparisons between a language model and its three fixed rivals are unaffected by this bug, since every agent at one table shares one draw of the market regardless of which seed produced it; what is not guaranteed is that the exact day-by-day sequence would reproduce bit-for-bit from the base seed alone before this fix.
Second, partway through data collection we introduced an explicit \texttt{think=false} request to Ollama for the DeepSeek-R1 family specifically, after observing call latencies far more variable than any other family's; this had no effect on Llama or Qwen (confirmed directly: identical response behavior with and without the flag) and was applied starting from that family's shipper-role runs onward, so DeepSeek-R1's carrier-role data reflects the model's default reasoning behavior while its shipper-role data reflects the explicit request to suppress it.
Comparing the two directly (Table~\ref{tab:llm-compliance}) shows this request did \emph{not} reliably reduce latency at any DeepSeek-R1 scale -- mean latency is flat-to-slightly-worse between carrier and shipper for all three DeepSeek-R1 sizes -- so we report both halves of the dataset honestly rather than claim a fix that the data does not support, and the headline finding below (Tables~\ref{tab:llm-carrier}-\ref{tab:llm-shipper}) holds identically in the unaffected carrier-role data alone.

\begin{table}[H]
\centering
\caption{Carrier-side LLM results: match rate and bootstrap mean margin (95\% CI), by tier and family, against the three fixed same-side rivals in the same markets.}
\label{tab:llm-carrier}
\small
\begin{tabular}{llrrr}
\toprule
Tier & Model & match & mean margin & 95\% CI \\
\midrule
\multicolumn{5}{l}{\textit{fixed rivals (pooled across all nine LLM markets)}} \\
 & ML2  & 1.00 & 68.6  & [66.5, 70.8] \\
 & OPT2 & 1.00 & 63.9  & [61.5, 66.3] \\
 & R    & 1.00 & $-$60.9 & [$-$68.8, $-$53.2] \\
\midrule
tiny   & llama3.2:1b      & 0.38 & $-$18.1  & [$-$33.2, $-$3.8] \\
tiny   & qwen2.5:1.5b     & 0.80 & $-$63.6  & [$-$88.1, $-$40.7] \\
tiny   & deepseek-r1:1.5b & 0.90 & $-$64.6  & [$-$89.7, $-$39.0] \\
small  & llama3.1:8b      & 1.00 & $-$113.0 & [$-$134.1, $-$90.7] \\
small  & qwen2.5:7b       & 1.00 & $-$136.0 & [$-$157.4, $-$112.5] \\
small  & deepseek-r1:7b   & 0.97 & $-$87.6  & [$-$114.3, $-$60.0] \\
medium & llama3.3:70b     & 0.82 & \textbf{+25.4}  & [+5.6, +43.2] \\
medium & qwen2.5:72b      & 1.00 & $-$20.9  & [$-$44.5, +0.9] \\
medium & deepseek-r1:70b  & \textbf{0.03} & $-$0.4   & [$-$1.3, 0.0] \\
\bottomrule
\end{tabular}
\end{table}

\begin{table}[H]
\centering
\caption{Shipper-side LLM results: match rate and bootstrap mean profit (95\% CI).}
\label{tab:llm-shipper}
\small
\begin{tabular}{llrrr}
\toprule
Tier & Model & match & mean profit & 95\% CI \\
\midrule
\multicolumn{5}{l}{\textit{fixed rivals (pooled across all nine LLM markets)}} \\
 & R    & 1.00 & 1163.1 & [1101.6, 1225.4] \\
 & OPT2 & 1.00 & 1148.8 & [1082.2, 1210.8] \\
 & ML2  & 0.99 & 1117.7 & [1055.4, 1180.2] \\
\midrule
tiny   & llama3.2:1b      & 0.50 & 2.7    & [0.0, 7.8] \\
tiny   & qwen2.5:1.5b     & 0.97 & 462.5  & [294.4, 652.3] \\
tiny   & deepseek-r1:1.5b & 0.95 & 379.8  & [224.6, 559.0] \\
small  & llama3.1:8b      & 1.00 & 548.6  & [390.0, 716.0] \\
small  & qwen2.5:7b       & 0.60 & 406.9  & [228.9, 610.6] \\
small  & deepseek-r1:7b   & 1.00 & 584.9  & [418.3, 756.8] \\
medium & llama3.3:70b     & 0.97 & \textbf{1264.1} & [1073.5, 1454.0] \\
medium & qwen2.5:72b      & 1.00 & 908.8  & [741.9, 1082.6] \\
medium & deepseek-r1:70b  & \textbf{0.03} & 3.8    & [0.0, 11.3] \\
\bottomrule
\end{tabular}
\end{table}

\begin{figure}[H]
\centering
\includegraphics[width=\textwidth]{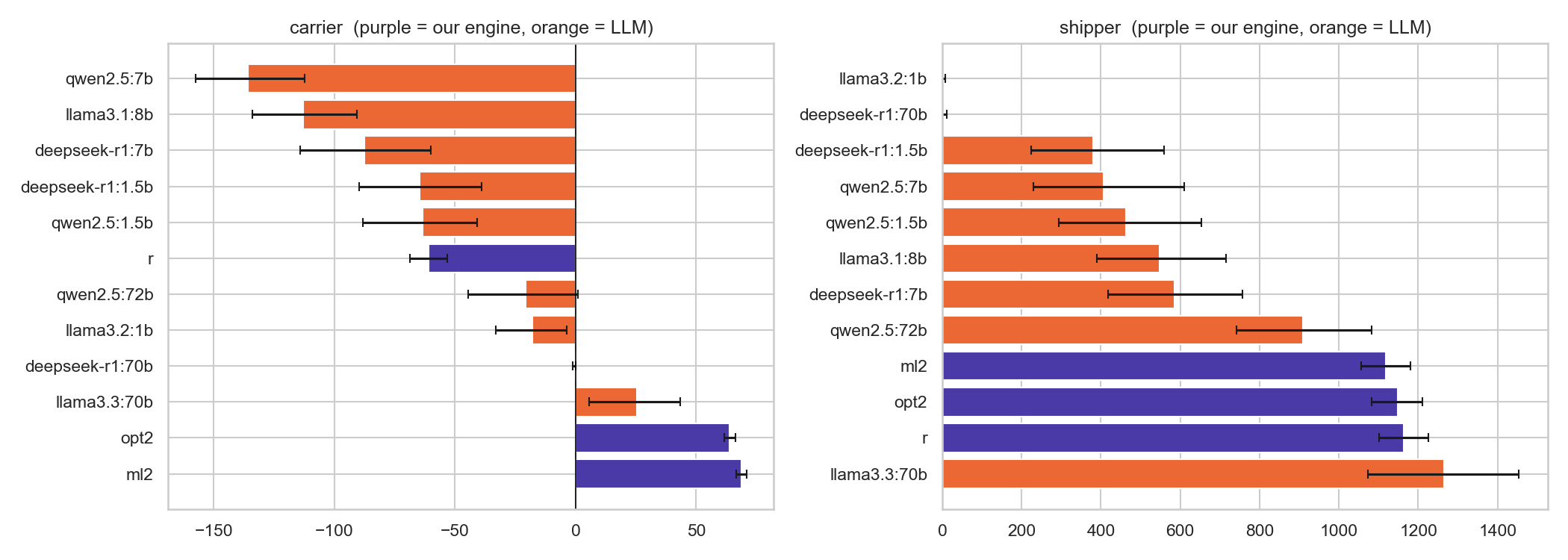}
\caption{Each language model's bootstrap mean outcome against its three fixed same-side rivals (purple), by role. Orange bars are the nine language models.}
\label{fig:llm-vs-motores}
\end{figure}

Across all $9\times2\times3=54$ pairwise comparisons between a language model and one of its three fixed rivals, 42 are significant losses for the language model, 4 are significant wins, and 8 are inconclusive after Holm-Bonferroni correction within each (model, role) family of three.
The pattern is sharply asymmetric by rival: against OPT2 and ML2 combined, language models lose 32 of 36 comparisons significantly and never win a single one (the remaining 4 are inconclusive, not reversed); against R, the weakest hand-designed architecture, the record is far more even -- 10 significant losses, 4 significant wins, 4 inconclusive.
Every one of the four significant wins against R is on the carrier side (\texttt{llama3.2:1b}, \texttt{llama3.3:70b}, \texttt{qwen2.5:72b}, and \texttt{deepseek-r1:70b}); on the shipper side, every language model loses to R significantly except \texttt{llama3.3:70b}, whose comparison is inconclusive with a positive point estimate ($+139.6$, $p_{\text{holm}}=0.80$).
Two of those four wins, however, come from models with very low carrier-side match rates (\texttt{llama3.2:1b} at 38\%, \texttt{deepseek-r1:70b} at just 3\%, Table~\ref{tab:llm-carrier}): since an unmatched day scores exactly $0$ (Section~\ref{sec:matching}) against R's own negative mean margin ($-$60.9, Table~\ref{tab:llm-carrier}), most of each model's advantage reflects how often it scored a forfeit's $0$ rather than consistently out-negotiating R; only \texttt{qwen2.5:72b} (100\% match) and \texttt{llama3.3:70b} (82\% match) win against R by actually closing deals at better terms.
\texttt{llama3.3:70b} is this comparison's standout performer by a wide margin: it is the only language model to beat the weakest rival on both roles (significantly on carrier, inconclusively on shipper), the only one whose comparisons against OPT2 and ML2 are not significant losses on the shipper side (point estimates of $+161.7$ and $+111.0$ from the underlying paired tests of Section~\ref{sec:stats}, not shown in a table), and it posts the single highest mean profit of all twelve shipper-side participants in this comparison, language models and fixed rivals alike ($1264.1$, ahead of R's $1163.1$, Table~\ref{tab:llm-shipper}).

\texttt{deepseek-r1:70b} shows the opposite pattern: a match rate of \textbf{2.5\% on both roles}, against 38-100\% for every other model in this comparison including its own 1.5B and 7B siblings (Tables~\ref{tab:llm-carrier}-\ref{tab:llm-shipper}).
Table~\ref{tab:llm-compliance} traces this to response latency: \texttt{deepseek-r1:70b}'s calls average 91.6-92.0s against a 90s timeout, essentially pinned at the ceiling on both roles regardless of the \texttt{think=false} request, with a first-try structured-output success rate of only 25\%/26\% (carrier/shipper), far below its own 1.5B (49-52\%) and 7B (55-88\%) siblings and at or below every Llama or Qwen model at any tier (26-98\%).
Unlike the engine tournament's match-rate collapse (Section~\ref{sec:results-collapse}), which traced to a training-time gap (no competitive pressure during training), this one traces to an inference-time constraint: \texttt{deepseek-r1:70b} is slower at this task than either its own smaller siblings or the same-tier Llama/Qwen models, and the explicit \texttt{think=false} request did not change that latency profile.
This constraint is specific to our hardware as much as to the model: on the single 24\,GB GPU of Section~\ref{sec:hardware}, all three 70B-class models partly run from CPU memory, and only \texttt{deepseek-r1:70b} was slow enough there to hit the 90-second timeout on nearly every call.
A machine able to hold the whole model in GPU memory, or a longer timeout, could change this result, so we report it as a deployment constraint of this setup, not as a property of the model family.

\begin{table}[H]
\centering
\caption{Structured-output compliance and latency, by tier, family, and role (DeepSeek-R1's \texttt{think} setting differs by role; see the methodological note above).}
\label{tab:llm-compliance}
\small
\resizebox{\textwidth}{!}{%
\begin{tabular}{llrrr}
\toprule
Tier & Model (role) & first-try-ok & mean latency (s) & p95 latency (s) \\
\midrule
tiny   & llama3.2:1b (carrier / shipper)      & 0.32 / 0.26 & 2.3 / 2.3   & 2.5 / 2.4 \\
tiny   & qwen2.5:1.5b (carrier / shipper)     & 0.45 / 0.58 & 2.3 / 2.3   & 2.4 / 2.4 \\
tiny   & deepseek-r1:1.5b (carrier / shipper) & 0.49 / 0.52 & 17.1 / 21.7 & 92.1 / 92.1 \\
small  & llama3.1:8b (carrier / shipper)      & 0.88 / 0.88 & 2.6 / 2.5   & 2.7 / 2.6 \\
small  & qwen2.5:7b (carrier / shipper)       & 0.55 / 0.81 & 2.4 / 2.4   & 2.5 / 2.6 \\
small  & deepseek-r1:7b (carrier / shipper)   & 0.55 / 0.88 & 9.0 / 10.6  & 15.0 / 22.9 \\
medium & llama3.3:70b (carrier / shipper)     & 0.68 / 0.98 & 29.3 / 30.2 & 37.8 / 38.1 \\
medium & qwen2.5:72b (carrier / shipper)      & 0.78 / 0.93 & 41.6 / 31.0 & 66.6 / 38.0 \\
medium & \textbf{deepseek-r1:70b} (carrier / shipper)  & \textbf{0.25 / 0.26} & \textbf{91.6 / 92.0} & \textbf{92.1 / 92.1} \\
\bottomrule
\end{tabular}%
}
\end{table}

\begin{figure}[H]
\centering
\includegraphics[width=\textwidth]{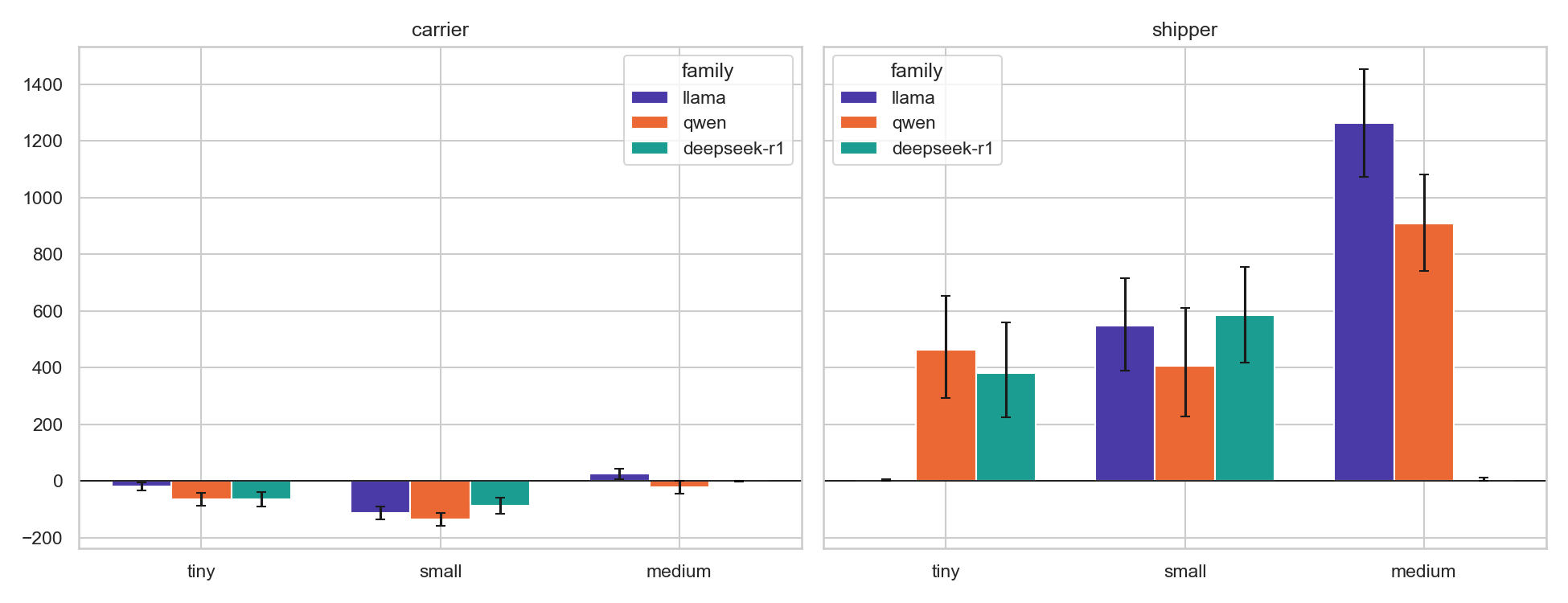}
\caption{Bootstrap mean outcome by tier and family, language models only. Error bars are 95\% bootstrap CIs.}
\label{fig:llm-tier-family}
\end{figure}

Pooling across family, scale does not help uniformly: on the carrier side the tiny tier ($-48.7$) actually beats the small tier ($-112.6$), with the medium tier best of all ($+1.4$) -- a non-monotonic pattern driven by the same forfeit-inflation effect that follows mechanically from the scoring convention established in Section~\ref{sec:matching}, since a model that fails to produce a legal move scores exactly $0$ for that day rather than a large negative margin, and the tiny tier's lower match rates (Tables~\ref{tab:llm-carrier}) mechanically pull its mean toward zero.
On the shipper side, by contrast, the pattern is monotonic -- tiny ($281.7$) $<$ small ($511.7$) $<$ medium ($721.0$) -- but this is not a clean scaling effect either: it is driven almost entirely by \texttt{llama3.3:70b}'s standout result averaged against \texttt{deepseek-r1:70b}'s near-total collapse, two same-tier models roughly $1260$ profit units apart (Table~\ref{tab:llm-shipper}).
We report both the pooled tier summary and this per-family decomposition explicitly (Figure~\ref{fig:llm-tier-family}; full figures in the released \texttt{tier\_summary.csv} and \texttt{tier\_family\_summary.csv}) rather than the tier average alone, since the average on its own would misrepresent both the carrier-side forfeit artifact and the shipper-side family divergence as if they were uniform scaling effects.
We report this comparison as a direct extension of the same experimental and statistical protocol used for the nine architectures above, so that language-model-based agents can be read on the same ranking and the same corrected-significance terms as every hand-designed, search-based, imitation-learned, and reinforcement-learned architecture in this paper.

\subsection{Negotiation trajectories: qualitative case studies}
\label{sec:results-llm-trajectories}

The aggregate statistics of Section~\ref{sec:results-llm} summarize $9\times2\times40=720$ negotiation-days into bootstrap means and pairwise tests; this subsection looks directly at four individual historical days, reconstructed from the per-call logs already saved during data collection, to give the aggregate numbers qualitative texture.
No negotiation was re-run and no model was called again to produce these examples: every value and every action below is exactly what the model emitted at the time, read back from the stored JSON response for that call.
One limitation of this reconstruction is unavoidable: the call log records only the language model's own emitted action each turn, never its counterpart's offer (which was visible to the model inside its prompt but was not separately persisted), so the figures below show one side's concession path only, not a two-sided price-convergence plot; a day index was also not logged directly, so individual historical days within one table were recovered by detecting resets in the turn/attempt sequence (a non-decreasing run of turns is one day; a drop back to an earlier turn marks the start of the next).

\begin{figure}[H]
\centering
\includegraphics[width=\textwidth]{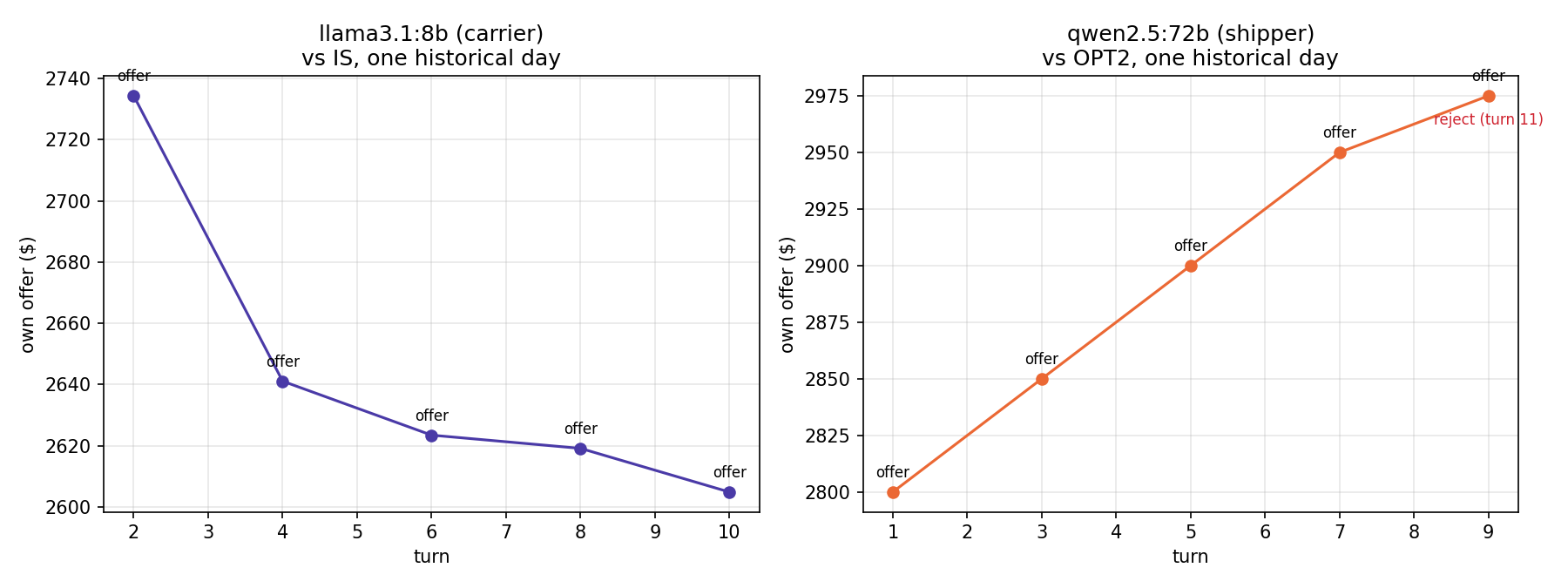}
\caption{Two ordinary negotiations, reconstructed from saved per-call logs. Left: \texttt{llama3.1:8b} (carrier) vs.\ IS, conceding from \$2734.31 to \$2604.91 over five turns. Right: \texttt{qwen2.5:72b} (shipper) vs.\ OPT2, conceding from \$2800.00 to \$2975.00 over five offers before being rejected at the final turn. Only the language model's own emitted offers are shown; the counterpart's offers were not separately persisted during data collection.}
\label{fig:llm-trajectory-clean}
\end{figure}

Figure~\ref{fig:llm-trajectory-clean} shows two ordinary, rule-compliant negotiations, one closed and one not.
On the left, \texttt{llama3.1:8b} as carrier concedes downward across five turns against IS, from an opening offer of \$2734.31 to \$2604.91 -- a smooth, monotonically non-increasing path consistent with the hard-band rule every architecture in this tournament obeys (Section~\ref{sec:protocol}).
On the right, \texttt{qwen2.5:72b} as shipper concedes upward from \$2800.00 to \$2975.00 across five offers against OPT2, and is still rejected at the final turn -- a reminder that a reasonable-looking, rule-compliant concession path does not by itself guarantee a deal, consistent with this model's non-perfect match rate reported in Table~\ref{tab:llm-shipper}.

\begin{figure}[H]
\centering
\includegraphics[width=0.75\textwidth]{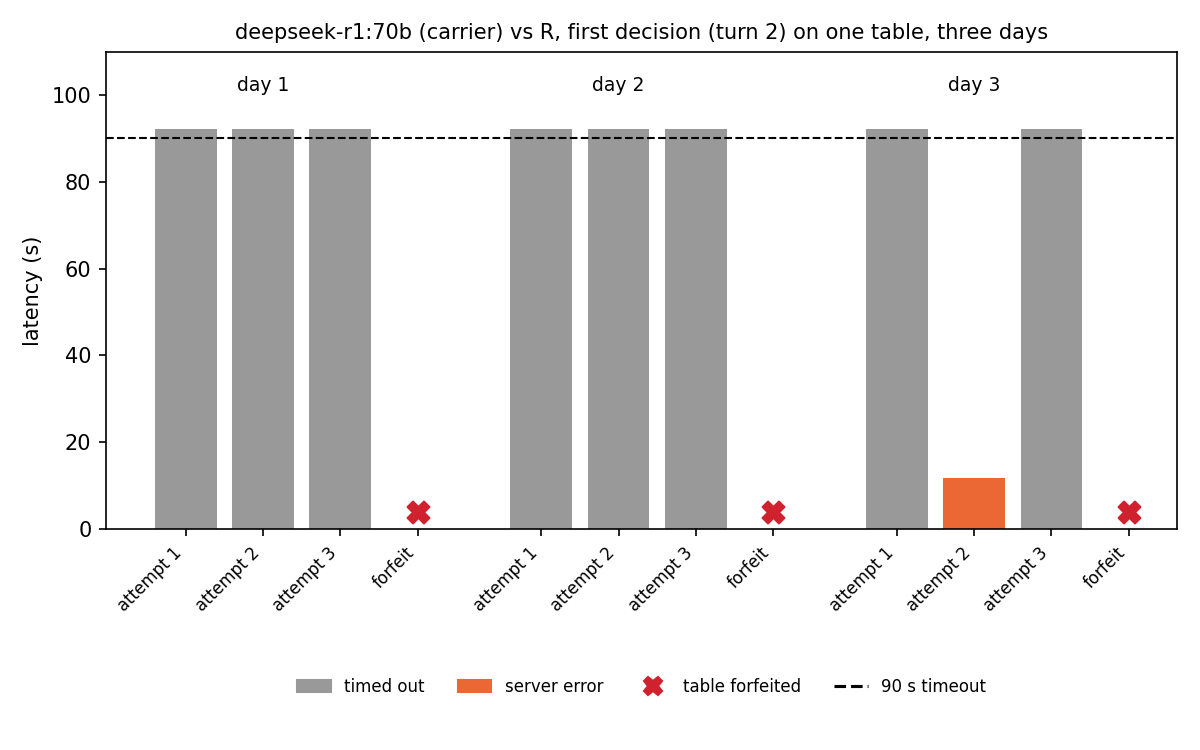}
\caption{\texttt{deepseek-r1:70b} (carrier) vs.\ R, the model's first decision (turn 2) on one table on three successive logged days. Each bar is one call attempt: grey bars timed out near the 90\,s request ceiling (dashed line), the orange bar failed earlier with a server-side error, and the red cross after each group of three attempts marks the table's forfeit once the retry budget was exhausted. No attempt produced a usable response.}
\label{fig:llm-trajectory-timeout}
\end{figure}

Figure~\ref{fig:llm-trajectory-timeout} shows \texttt{deepseek-r1:70b}'s collapse directly rather than only as a summary statistic: on three successive logged days, the model's first decision on one carrier table (turn 2) used up its whole retry budget -- three call attempts, the original plus \texttt{max\_retries=2} -- and the table was forfeited each time.
Eight of the nine attempts ended at essentially the same $\sim$92\,s wall-clock latency against the 90\,s request timeout, and the ninth failed earlier, after 11.7\,s, with a server-side error from the local Ollama server; no attempt produced a usable response, the same pattern Table~\ref{tab:llm-compliance} reports in aggregate.

Table~\ref{tab:llm-trajectory-struggle} shows a different kind of failure: in one historical day, \texttt{deepseek-r1:7b} (shipper, vs.\ IS) repeatedly confuses itself about its own action schema rather than failing on latency or format.
At turn 3 it twice emits \texttt{accept} with a non-null \texttt{amount} -- contradicting the schema's own stated rule that \texttt{amount} must be null for any action other than \texttt{offer} -- before a third attempt pivots to a legal \texttt{offer} that is used; at turn 7 it emits a schema-legal \texttt{accept} (\texttt{amount=null}) that is still not accepted (nothing on the table yet satisfied its own band), again pivoting to \texttt{offer} on the retry; at turn 9, shown in full below, it repeats the identical non-null-\texttt{amount} \texttt{accept} three consecutive times, exhausts its retry budget, and the table forfeits.
This is the same model whose first-try structured-output compliance (Table~\ref{tab:llm-compliance}) trails every other small-tier model on the shipper side; this example shows directly what that lower compliance rate consists of in one specific instance.

\begin{table}[H]
\centering
\caption{\texttt{deepseek-r1:7b} (shipper, vs.\ IS), turn 9 of one historical day: three consecutive illegal \texttt{accept} attempts before the table forfeits. Reconstructed from the saved call log; no negotiation was re-run.}
\label{tab:llm-trajectory-struggle}
\small
\begin{tabular}{llll}
\toprule
Attempt & Action & Amount & Outcome \\
\midrule
0 & accept & 3022.19 & retried: \texttt{amount} must be null for \texttt{accept} \\
1 & accept & 3022.19 & retried: same schema violation repeated \\
2 & accept & 3022.19 & retried: same schema violation repeated \\
-- & forfeit & -- & retry budget exhausted (\texttt{exhausted\_retries}) \\
\bottomrule
\end{tabular}
\end{table}

\section{Discussion}
\label{sec:discussion}

Three results now stand together, each visible only because this tournament tests every architecture, hand-designed or learned, against genuine same-side competition under the same statistical protocol, rather than treating any one family's agents as a special case.

The first is a training-time gap: a reinforcement-learned policy trained without competitive pressure collapses in match rate once rival agents compete with it under scarcity (Section~\ref{sec:results-collapse}).

The second is a train/deployment feature mismatch: a policy trained \emph{with} competitive pressure avoids that collapse, but is undone on realized payoff because the feature it depends on is not presented identically in training and at deployment (Section~\ref{sec:results-rl2gap}).

The third is an inference-time constraint specific to one model on our hardware: a large language model prompted with the same rules and the same opponents loses to this tournament's two strongest engines in the overwhelming majority of comparisons (32 of 36 against OPT2/ML2), with the one partial exception (\texttt{llama3.3:70b}) still never reaching significance against either, and the largest model in its own family (\texttt{deepseek-r1:70b}) failing for a third, distinct reason: a latency ceiling, on the single-GPU workstation of Section~\ref{sec:hardware}, that it hit on nearly every call, unlike its own smaller siblings and its same-tier rivals, and that did not respond to disabling its explicit reasoning step.

On the carrier side of a balanced market, this tournament's best-performing architectures are not the simplest rule-based ones, nor the most expensive search-based one, but the two families that compute or learn an opponent-aware offer directly: continuous expected-utility optimization with an explicitly fitted opponent belief (OPT2) and offer imitation learned from self-play together with a separately learned accept classifier (ML2); on the shipper side of the same balanced market, by contrast, OPT2 and ML2 rank seventh and eighth of nine (Table~\ref{tab:shipper}), so this opponent-awareness advantage does not generalize across roles.
Sophistication in how an offer is computed is not, by itself, sufficient to guarantee a good outcome once competition from rival agents and scarcity are introduced: Section~\ref{sec:results-collapse}'s match-rate collapse shows that an architecture (RL) trained to a plausible-looking reward in an isolated environment can fail specifically and severely once deployed against rival agents, in exactly the market condition designed to apply that competitive pressure, while Section~\ref{sec:results-rl2gap}'s finding shows that training against genuine competitive pressure (RL2) is not sufficient either, if a feature the resulting policy depends on is not presented identically between training and deployment.
Both findings would be invisible to an evaluation protocol built only from isolated one-on-one duels or a single fixed market condition, which is the gap this paper's many-sided, scarcity-varying, statistically corrected tournament was built to close.

\subsection{Comparison with prior automated-negotiation and LLM-negotiation benchmarks}
\label{sec:discussion-sota}

Section~\ref{sec:related} positioned this tournament against three prior literatures; the results above let that comparison be made concretely rather than only structurally.

Against the GENIUS/ANAC ecosystem \citep{lin2014genius}: \citet{dejonge2022micro} showed that a strategy with no opponent modeling or learning at all (MiCRO) beats the strongest submitted ANAC agents, because the classic linear-additive domains make opponent-aware sophistication unnecessary for winning a single isolated duel.
This tournament's own ranking (Section~\ref{sec:results-ranking}) shows the opposite pattern once a populated market and real matching are added: the top of the carrier ranking is held by architectures that either learn an offer from data (the imitation-learned ML2, the reinforcement-learned RL) or compute one by optimizing against a modeled counterpart threshold (OPT and, with an explicitly fitted belief, OPT2), while the simplest architectures with neither property (R and, to a lesser extent, RV) occupy the bottom of the carrier ranking in every market condition (Table~\ref{tab:carrier}).
The two results do not contradict each other -- MiCRO's advantage is specific to ANAC's classic domains' utility-function structure, and this tournament deliberately keeps that structure simple while making the market structure itself (scarcity, rival density, a real matching stage) the source of difficulty instead -- but together they show that whether opponent-aware sophistication pays off depends on which axis of the environment is actually hard, a question no single benchmark answers for both axes at once.

Against learned negotiation dialogue \citep{lewis2017,he2018}: those systems are trained and evaluated in the same isolated-bilateral setting their architecture assumes, so their own evaluation protocol has no equivalent of this tournament's match-rate collapse (Section~\ref{sec:results-collapse}).
Our RL architecture inherits the identical training assumption -- reward from closing a deal in an isolated environment -- and this tournament is, to our knowledge, the first time that specific assumption's cost under competition from rival agents and scarcity has been measured directly, rather than left untested by an evaluation protocol that never introduces rival agents.

Against the 2023--2026 LLM-negotiation benchmarks \citep{abdelnabi2023,pollo2026reproduction,termsbench2026,oh2026merit}: our own results echo their shared finding that a single success-rate or deal-rate number undersells what is actually happening.
\texttt{llama3.3:70b} posts the single highest raw shipper-side profit of all twelve participants in its comparison (Table~\ref{tab:llm-shipper}) without its comparisons against OPT2/ML2 reaching significance, and \texttt{deepseek-r1:70b}'s near-total match-rate collapse (Section~\ref{sec:results-llm}) is a specific, diagnosable inference-time latency artifact of our hardware and 90-second timeout, not a general reasoning failure of that model family -- exactly the kind of decomposition \citet{termsbench2026} and \citet{oh2026merit} argue deal-rate alone would hide, and consistent with \citet{pollo2026reproduction}'s caution that a single LLM-negotiation ranking is rarely the full story.
What none of these benchmarks include is a rival roster of hand-designed and learned (non-LLM) architectures inside the same populated, scarcity-varying market: our finding that language models lose to this tournament's two strongest engines in the large majority of comparisons (Section~\ref{sec:results-llm}) is, to our knowledge, the first time that specific comparison -- an LLM negotiator against a genuinely competitive non-LLM field, under real matching and scarcity -- has been measured directly, echoing in bilateral price negotiation the same direct-policy-versus-model-and-search pattern \citet{lehrach2025} report in general game-playing.

This paper's contribution, stated against that background, is not a new negotiation domain, a new learned architecture, or a new LLM-evaluation metric considered alone: each of those already exists in one of the three literatures above.
It is the combination of (i) a populated, many-sided market cleared by a real matching mechanism rather than one isolated bilateral exchange, (ii) scarcity varied systematically across every possible exclusion pair rather than fixed at one condition, and (iii) one shared, statistically corrected protocol applied uniformly to rule-based, search-based, optimization-based, imitation-learned, reinforcement-learned, and large-language-model-based agents at once, letting all six families be ranked against each other directly rather than compared only within their own sub-literature's separate benchmark, with (iv) one normalized negotiation distance (Eq.~\eqref{eq:valuebar}) as the common scale on which eight of the nine engine architectures value, optimize, or learn their offers.

This study has several limitations.

The market model (Section~\ref{sec:market}) is a synthetic, mean-reverting price process over six fixed routes, calibrated for internal consistency rather than fit to real freight-market data, so the absolute magnitudes reported here should not be read as predictions for a real market; the relative comparison between architectures under the same synthetic market is the result this paper makes.

Every architecture ran frozen, with no within-tournament parameter adaptation, so this paper measures nine fixed policies against each other and not an open-ended co-adaptation process; each architecture's hyperparameters (concession rates, particle counts, optimizer step counts, and so on) were set once rather than tuned per architecture or per market condition, so a reported ranking reflects one hyperparameter setting per architecture, not necessarily each architecture's best achievable performance.

The paired significance tests themselves have three unaddressed limitations: the underlying log-space Ornstein-Uhlenbeck price process (Section~\ref{sec:market}) generates serially correlated draws across consecutive negotiation indices rather than independent and identically distributed ones, and the paired Wilcoxon signed-rank test's standard assumptions do not account for that correlation; we do not report which zero-difference handling method (Wilcoxon's, Pratt's, or zero-split) was used on the rare pairs whose per-day outcomes tied exactly; and we report only Holm-Bonferroni-corrected significance for each of the 36 comparisons per condition and role, not a standardized effect size alongside it.

We deliberately report RL2 as actually deployed, with its confirmed train/deployment feature mismatch left unpatched, rather than silently correcting it before evaluation, since the gap between an architecture's potential and its actually-measured performance is itself part of what this tournament is designed to surface; a corrected RL2, retrained with real portfolio signal available during training as well as at deployment, is a direct next step rather than a result reported here.
A reward-design alternative to that occupancy-feature explanation does not hold up under direct inspection of the training code: RL2's reward already includes an explicit margin-based term identical in form to RL's own terminal reward (the same bounded value-bar function of the closing price) whenever its table closes and it wins the match, conditioned on winning rather than omitted outright, so a training signal indifferent to realized margin is not a plausible competing explanation for its low realized payoff.

The language-model comparison adds its own limitations beyond those above: each model was prompted once, with one fixed system prompt and one fixed disclosure policy, so these results characterize that specific prompting choice rather than each model's best achievable performance under prompt engineering; the three fixed same-side rivals (R, OPT2, ML2) were chosen to span this tournament's weak/strong/learned range rather than to include every architecture, so a language model's record against the other six architectures is not directly measured; and, as reported in Section~\ref{sec:results-llm}, the DeepSeek-R1 family's carrier-role and shipper-role data were collected under different explicit reasoning settings after a mid-study methodological change, which we verified does not alter this family's headline latency-collapse finding but which is disclosed rather than hidden; and every latency figure, including that collapse, was measured on one workstation whose 24\,GB GPU cannot hold any of the 70B-class models whole (Section~\ref{sec:hardware}), so different hardware or a longer timeout could change the latency-driven results.

The trajectory case studies of Section~\ref{sec:results-llm-trajectories} are similarly limited by construction: the call log used to reconstruct them never persisted a counterpart's offer, only the language model's own emitted action each turn, so those figures and that table show one side's concession path, not a two-sided price-convergence plot.

Future work includes retraining RL2 with real portfolio signal available during training, a hyperparameter sweep across architectures and market conditions, extending the scarcity design to ratios other than seven-of-nine, calibrating the market model against real freight-brokering data, and persisting the counterpart's offer directly during future data collection so that a trajectory analysis can be two-sided rather than reconstructed from one side's logs alone.

\section{Conclusion}
\label{sec:conclusion}

We built nine negotiation-engine architectures on a shared interface and compared them, for the first time in this project, inside a single many-sided competitive tournament rather than isolated duels, under three market conditions spanning a balanced market and two systematically-rotated scarcity conditions, with every pairwise comparison corrected for the resulting 36 simultaneous tests.
Two results from that tournament would not have been visible to a narrower evaluation: a reinforcement-learned policy trained without competitive pressure suffers a severe, condition-specific match-rate collapse once rival agents compete with it under scarcity, and a reinforcement-learned policy trained \emph{with} competitive pressure avoids that collapse but is instead undone, on realized payoff, by a confirmed mismatch between a feature's value during training and during deployment.

Extending the same protocol to nine language models tested them head-to-head against three of this tournament's nine architectures (R, OPT2, ML2) rather than leaving that class of agent untested: against the two strongest of those three, OPT2 and ML2, language models lost significantly in 32 of 36 comparisons and won none, confirming that the architectural advantages identified above are not artifacts of excluding a currently-popular class of agent, while the one partial exception (\texttt{llama3.3:70b}) and the one severe, inference-time collapse on our hardware (\texttt{deepseek-r1:70b}) show that a language model's negotiating competence in this setting depends on which specific model and tier is deployed at least as much as on the fact that it is a language model at all.
Together these results argue that a negotiation architecture's quality cannot be fully assessed from isolated duels or a single fixed market condition alone, and that a many-sided, scarcity-varying, statistically corrected tournament is a practical way to surface the specific failure modes -- competitive collapse, and train/deployment gaps -- that matter once such an architecture is actually deployed into a market with rival agents.

\section*{Declarations}
\label{sec:declarations}

\textbf{Funding.} This research was conducted as part of the authors' work in the AI Lab at Freight Technologies Inc. and received no external grant funding.

\textbf{Competing interests.} Eduardo de Avila-Armenta, Ra\'ul Alejandro Aquino-Ibarra, Jorge Octavio Nicol\'as-D\'iaz, and Umberto Leon-Dominguez are employed by Freight Technologies Inc., a company that develops automated freight-brokering and negotiation technology directly related to the subject matter of this paper. This relationship is disclosed as a potential competing interest.

\section*{Data and Code Availability}
\label{sec:availability}
The simulation engine, all nine negotiation architectures, the LLM validation harness, and the raw result logs analyzed in this paper (including the per-call logs reconstructed in Section~\ref{sec:results-llm-trajectories}, \texttt{data/llm\_validation/*\_calls.jsonl}, and the summary files \texttt{tier\_summary.csv} and \texttt{tier\_family\_summary.csv} referenced in Section~\ref{sec:results-llm}, and \texttt{data/validation/pct\_outcomes.csv} with the percentage columns of Tables~\ref{tab:carrier}-\ref{tab:shipper}, computed from the stored results by \texttt{scripts/\_compu-\\te\_pct\_outcomes.py}) are available at \href{https://github.com/AILabFr8tech/wehaveadeal}{github.com/AILabFr8tech/wehaveadeal} (the paths above are relative to its \texttt{paper/} folder) and archived on Zenodo at \href{https://doi.org/10.5281/zenodo.24122038}{doi:10.5281/zenodo.24122038}.

\bibliographystyle{plainnat}
\bibliography{refs}

\end{document}